\documentclass[journal]{IEEEtran}
\IEEEoverridecommandlockouts

\usepackage{amsmath,amsfonts}
\usepackage{algorithmic}
\usepackage{algorithm}
\usepackage{array}
\usepackage{subfigure}
\usepackage{textcomp}
\usepackage[table]{xcolor}
\usepackage{tabularx}
\usepackage{booktabs}
\usepackage{threeparttable}
\usepackage{makecell}
\usepackage{multirow}
\usepackage{xurl}
\usepackage[colorlinks=true, allcolors=black]{hyperref}
\usepackage{stfloats}
\usepackage{url}
\usepackage{verbatim}
\usepackage{graphicx}
\usepackage{cite}
\usepackage{adjustbox}
\definecolor{mygreen}{rgb}{0.13, 0.55, 0.13}

\usepackage{acro}
\usepackage{enumitem}

\DeclareAcronym{ai}{
  short = AI,
  long  = Artificial Intelligence,
}
\DeclareAcronym{adas}{
  short = ADAS,
  long  = Advanced Driver Assistance Systems,
}
\DeclareAcronym{ann}{
  short = ANN,
  long  = Artificial Neural Networks,
}
\DeclareAcronym{arima}{
  short = ARIMA,
  long  = Auto-Regressive Integrated Moving Average,
}
\DeclareAcronym{cnn}{
  short = CNN,
  long  = Convolutional Neural Networks,
}
\DeclareAcronym{dt}{
  short = DT,
  long  = Decision Tree,
}
\DeclareAcronym{gcn}{
  short = GCN,
  long  = Graph Convolutional Network,
}
\DeclareAcronym{gnn}{
  short = GNN,
  long  = Graph Neural Network,
}
\DeclareAcronym{gru}{
  short = GRU,
  long  = Gated Recurrent Unit,
}
\DeclareAcronym{knn}{
  short = KNN,
  long  = K-Nearest Neighbours,
}
\DeclareAcronym{lgbm}{
  short = LGBM,
  long  = Light Gradient-Boosting Machine,
}
\DeclareAcronym{lr}{
  short = LR,
  long  = Linear Regression,
}
\DeclareAcronym{lstm}{
  short = LSTM,
  long  = Long Short-Term Memory,
}
\DeclareAcronym{ml}{
  short = ML,
  long  = Machine Learning,
}
\DeclareAcronym{mlp}{
  short = MLP,
  long  = Multi-Layer Perceptron,
}
\DeclareAcronym{mlr}{
  short = MLR,
  long  = Multiple Linear Regression,
}
\DeclareAcronym{mms}{
  short = MMS,
  long  = Micromobility System,
}
\DeclareAcronym{prisma}{
  short = PRISMA,
  long  = Preferred Reporting Items for Systematic Reviews and Meta-Analyses,
}
\DeclareAcronym{rf}{
  short = RF,
  long  = Random Forest,
}
\DeclareAcronym{rnn}{
  short = RNN,
  long  = Recurrent Neural Networks,
}
\DeclareAcronym{stgcn}{
  short = STGCN,
  long  = Spatio-Temporal Graph Convolutional Network,
}
\DeclareAcronym{svm}{
  short = SVM,
  long  = Support Vector Machines,
}
\DeclareAcronym{xgb}{
  short = XGB,
  long  = Extreme Gradient Boosting,
}

\acsetup{
  list/template = description
}
\setlist[description]{align=left, leftmargin=2cm, style=nextline}

\begin{document}
\title{Machine Learning in Micromobility: A Systematic Review of Datasets, Techniques, and Applications}


\author{Sen Yan, \IEEEmembership{Graduate Student Member, IEEE}, Chinmaya Kaundanya, \\ Noel E. O’Connor, \IEEEmembership{\textcolor{black}{Senior Member},  IEEE}, Suzanne Little, and Mingming Liu, \IEEEmembership{Senior Member, IEEE}
    
    \thanks{This publication has emanated from research conducted with the financial support of Taighde Éireann — Research Ireland under Grant number \textit{21/FFP-P/10266} and \textit{SFI/12/RC/2289\_P2}. (\textit{Corresponding author: Suzanne Little.)}}
    
    \thanks{S. Yan, N. O'Connor, and M. Liu are with the School of Electronic Engineering and Insight Research Ireland Centre for Data Analytics at Dublin City University, Dublin 9, Ireland. (e-mail: sen.yan@dcu.ie; noel.oconnor@dcu.ie; mingming.liu@dcu.ie).}

    \thanks{C. Kaundanya and S. Little are with the School of Computing and Insight Research Ireland Centre for Data Analytics at Dublin City University, Dublin 9, Ireland. (e-mail: chinmaya.kaundanya3@mail.dcu.ie; suzanne.little@dcu.ie).}}



\maketitle

\begin{abstract}
Micromobility systems, which include lightweight and low-speed vehicles such as bicycles, e-bikes, and e-scooters, have become an important part of urban transportation, with the potential to address traffic congestion, air pollution, and high transportation costs. Successful utilisation of micromobility requires optimisation of complex systems for operational efficiency, environmental impact mitigation, and user safety. Machine Learning (ML) methods are increasingly used to support prediction and decision-making in these systems. Existing reviews differ in their coverage of micromobility datasets, ML methods, and application settings. This review synthesises 95 studies from January 2019 to July 2024, providing a structured review of datasets, ML techniques, and their applications in micromobility. Specifically, we collect and analyse micromobility-related datasets in terms of their spatial, temporal, and feature-based characteristics. We also review ML models applied in micromobility, discussing their principles, strengths, limitations, and application settings. 
Furthermore, we examine applications including demand prediction, energy management, and safety, focusing on their evaluation conditions and practical implications. Finally, we identify research challenges and directions related to generalisation, fairness and equity, privacy, public transport integration, model deployment, and operational evaluation.
\end{abstract}


\printacronyms[
    heading=section*, 
    name=List of Abbreviations \& Acronyms, 
]
\section{Introduction}
\label{sec: intro}

\subsection{Micromobility in Urban Transportation}


\color{black}

As cities become more densely populated, demand for efficient and environmentally friendly transport continues to grow. The term ``micromobility'' has been defined in various ways. In this study, we adopt a working definition consistent with recent conceptualisations \cite{ITF2020, Oeschger2020, Abduljabbar2021}, characterising micromobility as lightweight ($<$350 kg), low-speed ($<$45 km/h) vehicles designed for short-distance urban travel, including bicycles, e-bikes, and e-scooters. Its growth has been driven by efforts to address traffic congestion, air pollution, and transportation costs \cite{Khalil2021, Yan2023_2, Jaber2023}. By providing last-mile connectivity, reducing dependence on private cars, and \textcolor{black}{
improving access to conventional public transport}, micromobility has become an important component of urban mobility \cite{Yan2022, VegaGonzalo2024, Liu2022}. Governments and policymakers are therefore increasingly incorporating micromobility into sustainable transport strategies\footnote{\url{https://www.gov.ie/en/publication/e62e0-electric-vehicle-policy-pathway}}\footnote{\url{https://highways.dot.gov/public-roads/spring-2021/02}}\footnote{\url{https://www.intertraffic.com/news/urban-mobility/china-mobility-revolution}}.

\color{black}

\subsection{Conventional Approaches for Micromobility}

Conventionally, \textcolor{black}{transport-related} problems have been addressed using established statistical, rule-based, optimisation, and engineering-driven approaches, e.g., GIS-based spatial analyses for demand characterisation \cite{Lopes2014}, gravity and network-based formulations for \textcolor{black}{spatial interaction analysis} \cite{Ceha1997, Liu2017}, and shortest-path or optimisation algorithms for route planning \cite{Fan2010, Guruji2016}. However, \textcolor{black}{the suitability of these methods depends on their assumptions and the available observations, including traffic, weather, and vehicle availability}. With the increase in high-resolution spatiotemporal data and improved computing and toolchains, \ac{ml} has increasingly \textcolor{black}{been applied to} modelling nonlinear micromobility dynamics and enabling real-time decision support.

\subsection{Machine Learning Methods for Micromobility}


\color{black}

\ac{ml} methods are increasingly used in micromobility to identify patterns from large-scale data and support prediction and adaptive decision-making. \textcolor{black}{
They can learn complex relationships from data and update predictions as new information becomes available, supporting responsive analysis of changing micromobility conditions}. Applications include demand prediction \cite{Guidon2019, Zhou2022, Saum2020} and energy management \cite{Ouf2023, Ina2022, Gioldasis2024}, with potential benefits for vehicle distribution and energy use. Overall, \ac{ml} can \textcolor{black}{
support more efficient, safe, and responsive micromobility systems.}

\color{black}

\subsection{Contributions \& Paper Structure}


\color{black}

\begin{figure}[ht]
    \vspace{-0.1in}
    \centering
    \includegraphics[width=\linewidth]{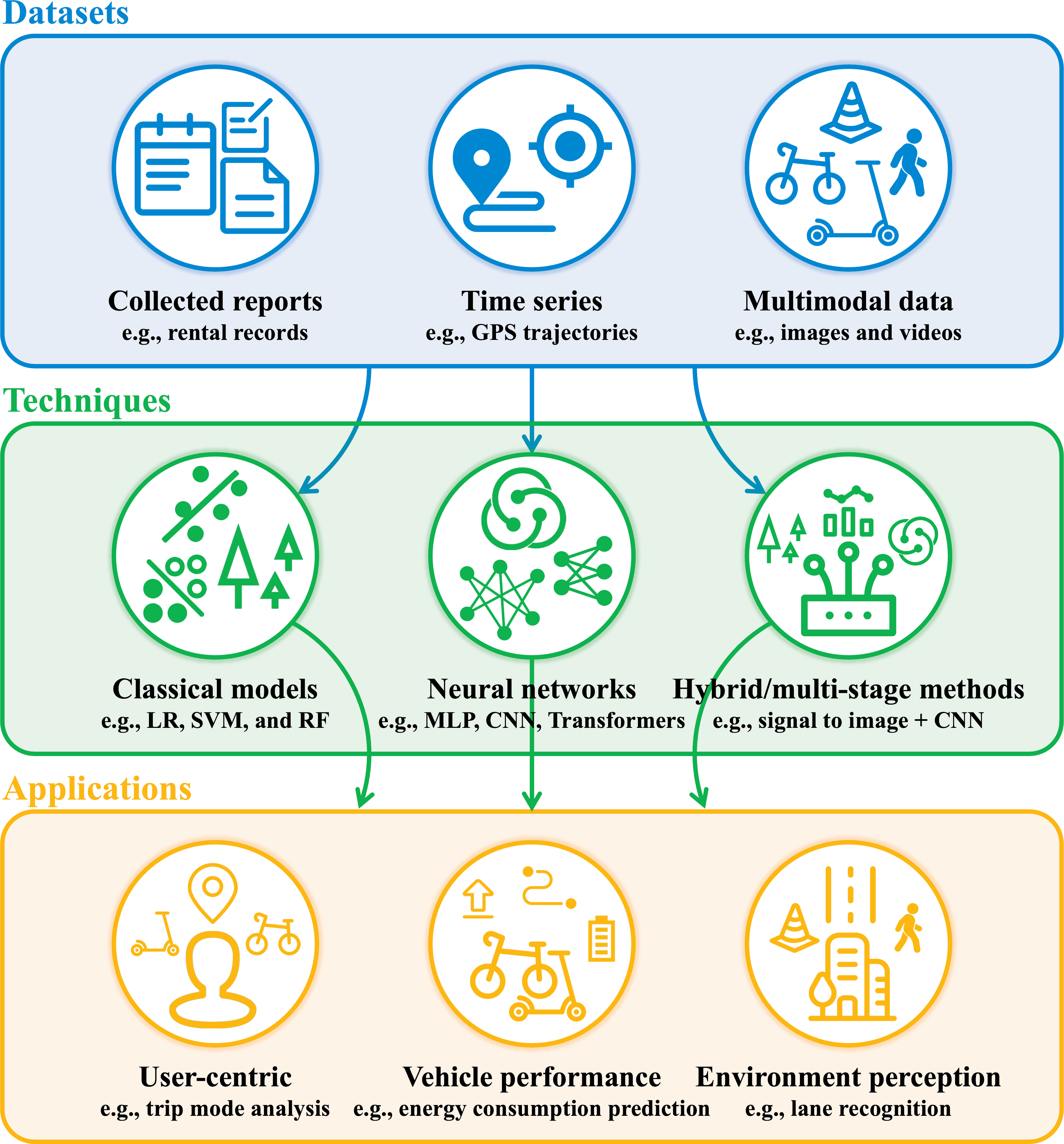}
    \caption{Research scope in this review paper.}
    \label{fig: abstract}
    \vspace{-0.1in}
\end{figure}

While some surveys have discussed \ac{ml} in broader mobility contexts \cite{Almukhalfi2024, Modi2021, Boukerche2020}, \textcolor{black}{
existing reviews differ in their coverage of micromobility-specific datasets, modelling assumptions, evaluation protocols, and application settings. This motivates our review of} existing datasets, \ac{ml} techniques, and their applications, as summarised in \autoref{fig: abstract}. The main contributions are:

\color{black}

\begin{itemize}
    \item A wide range of datasets relevant to micromobility are documented and analysed, providing insights into their spatial, temporal, and feature-based characteristics.
    \item Various \ac{ml} models applied in micromobility are introduced, categorised by their principles, advantages, and challenges, and discussed in their specific applications.
    \item Different applications of \ac{ml} methods are discussed in key areas of micromobility, such as demand prediction, route planning, and rider safety, \textcolor{black}{
    together with their evaluation conditions and practical implications}.
    \item Critical challenges in the current application of \ac{ml} to micromobility are identified, \textcolor{black}{
    including issues related to fairness and equity, privacy, public transport integration, model deployment, and operational evaluation,} along with future research directions to address these issues.
\end{itemize}

\color{black}

This paper is structured as follows: \textcolor{black}{
\autoref{sec: review} compares existing micromobility survey papers, followed by the review methodology in \autoref{sec: method}.} \autoref{sec: dataset}, \autoref{sec: tech}, and \autoref{sec: app} review datasets, \ac{ml} techniques, and applications, respectively. \autoref{sec: direction} discusses research directions and challenges, and \autoref{sec: conclusion} concludes the paper and outlines its limitations.

\color{black}


\section{\textcolor{black}{
Related Survey Papers}}
\label{sec: review}

\color{black}

This section positions this review relative to existing survey papers by comparing their scope and coverage, as summarised in \autoref{table: surveys}. Existing surveys span broader urban mobility systems \cite{Milakis2020, Teusch2023, Ahmed2022, Abduljabbar2021}, mode-specific topics such as bike sharing and e-scooters \cite{Albuquerque2021, Mitropoulos2023}, and specific technical objectives such as e-mobility energy management \cite{Yan2023}. However, micromobility is often treated as one component of broader transportation systems or examined through individual modes or applications. Dataset-related issues receive limited attention \cite{Teusch2023, Ahmed2022}, while cross-mode comparison and micromobility-specific demand analysis also remain limited \cite{Abduljabbar2021, Albuquerque2021, Mitropoulos2023}. \textcolor{black}{
Recent methodological surveys have also reviewed graph-based and spatiotemporal learning for transportation forecasting \cite{Jiang2022}, but their focus is primarily on graph-learning methodologies rather than the relationships among micromobility datasets, modelling assumptions, evaluation protocols, and application settings.}

\color{black}

\begin{table*}[ht]
    \caption{Summary of recent related survey papers.}
    \vspace{-0.1in}
    \label{table: surveys}
    \scriptsize
    \begin{tabularx}{\linewidth}{@{\extracolsep{\fill}}c X c c c c c c c c}
        \toprule
        \multirow{2}{*}[-2ex]{\textbf{Ref.}} & \multirow{2}{*}[-2ex]{\textbf{Brief Descriptions}} & \multicolumn{3}{c}{\textbf{Modalities}} & \multicolumn{5}{c}{\textbf{Topics}} \\
        \cmidrule(lr){3-5} \cmidrule(lr){6-10}
        & & bicycle & e-bike & e-scooter & datasets & \makecell{\ac{ml} \\methods} & \makecell{energy \\analysis} & \makecell{demand \\analysis} & safety \\
        \midrule
        \cite{Milakis2020} & \makecell[X]{The impact of micromobility on accessibility, air pollution, safety, physical activity, and subjective well-being.} & $\times$ & $\times$ & $\times$ & & & $\times$ & & $\times$ \\
        \midrule
        \cite{Albuquerque2021} & \makecell[X]{\ac{ml} techniques in bike-sharing systems, with a focus on usage pattern detection and demand prediction.} & $\times$ & & & & $\times$ & & $\times$ & \\
        \midrule
        \cite{Teusch2023} & \makecell[X]{\ac{ml} methods in shared transportation systems, with a focus on user analysis, demand analysis, dispatching, and infrastructure planning.} & $\times$ & $\times$ & $\times$ & $\times$ & $\times$ & $\times$ & \\
        \midrule
        \cite{Ahmed2022} & \makecell[X]{\ac{ml} methods applied to urban transportation, focused on public datasets, passenger localisation, transportation mode detection and pattern recognition, and generative mobility models.} & $\times$ & $\times$ & & $\times$ & $\times$ & & & \\
        \midrule
        \cite{Abduljabbar2021} & \makecell[X]{Journal articles in the Scopus database from 2000 to 2020, focusing on benefits, technology, policy, and behavioural model choices, with a very brief technical discussion.} & $\times$ & $\times$ & $\times$ & & $\times$ & & $\times$ & $\times$ \\
        \midrule
        \cite{Mitropoulos2023} & \makecell[X]{The properties of shared e-scooter systems and their impacts on the environment, user social, economic, safety, and transportation performance.} & & & $\times$ & & & $\times$ & & $\times$ \\
        \midrule
        \cite{Yan2023} & \makecell[X]{\ac{ml} approaches for energy management of e-mobility services.} & & $\times$ & $\times$ & & $\times$ & $\times$ & & \\
        \midrule
        \textcolor{black}{\cite{Jiang2022}} & \makecell[X]{\textcolor{black}{Graph neural networks for traffic forecasting, including bike-sharing and e-scooters; graph construction, spatio-temporal models, and open datasets.}} & \textcolor{black}{$\times$} & & \textcolor{black}{$\times$} & \textcolor{black}{$\times$} & \textcolor{black}{$\times$} & & \textcolor{black}{$\times$} & \\
        \midrule
        \textbf{\text{[$\ast$]}} & \makecell[X]{\textbf{\ac{ml} methods applied in shared micromobility, focused on datasets, applications, and methodologies using multimodal data.}} & \textbf{$\times$} & $\times$ & $\times$ & $\times$ & $\times$ & $\times$ & $\times$ & $\times$ \\
        \bottomrule
    \end{tabularx}
    \begin{tablenotes}\footnotesize
        \item The last item marked in \textbf{bold} represents our work conducted in this paper.
    \end{tablenotes}
    \vspace{-0.2in}
\end{table*}

\color{black}

Cross-comparison of the surveys in \autoref{table: surveys} reveals three recurring gaps in the existing literature: (1) broader reviews often examine micromobility within general mobility systems and emphasise accessibility, sustainability, or safety rather than micromobility-specific modelling issues \cite{Milakis2020, Abduljabbar2021, Teusch2023, Ahmed2022}; (2) technically focused surveys tend to concentrate on individual modes or tasks, such as bike-sharing demand prediction \cite{Albuquerque2021}, e-scooter systems \cite{Mitropoulos2023}, or e-mobility energy management \cite{Yan2023}, limiting cross-mode and cross-task comparison; and (3) dataset treatment remains uneven, with limited systematic comparison of data modalities, feature structures, and their suitability for different \ac{ml} tasks.

This review addresses these gaps by placing micromobility at the centre of analysis and synthesising datasets, \ac{ml} techniques, and application domains within a common framework. \textcolor{black}{
It also complements graph-learning surveys by focusing on how micromobility-specific data characteristics, modelling assumptions, validation practices, and application objectives are connected.}

\color{black}

\section{\textcolor{black}{
Review Methodology}}
\label{sec: method}

\color{black}


The \ac{prisma} framework was used to organise and report the literature identification and screening process \cite{Liberati2009}. \textcolor{black}{
The review procedure followed the detailed search-reporting guidance in \cite{Rethlefsen2021}.} The methodology is described through the study objectives, eligibility criteria, and search strategy.

\subsection{Study Objectives}

The formulation of study objectives is guided by the purpose of the literature review. The review was guided by three research questions:
\begin{itemize}
    \item What datasets related to micromobility have been published or used in existing research?
    \item Which \ac{ml} technologies have been employed in micromobility-relevant research papers?
    \item Which research problems have these \ac{ml} methods been applied to address?
\end{itemize}

These questions structure the review around micromobility datasets, \ac{ml} techniques, and application domains. Unlike broader surveys in which micromobility is typically considered alongside other transport modes, this review places micromobility at the centre of analysis.



\color{black}

\subsection{Eligibility Criteria}


\color{black}

This review focuses on \ac{ml} techniques applied to micromobility research rather than providing an exhaustive review of \ac{ml} algorithms. \textcolor{black}{The application scope follows the gaps identified in \autoref{sec: review}: demand and destination prediction, travel behaviour, energy management, route and fleet planning, and rider safety.} We included peer-reviewed studies applying \ac{ml} or fitted statistical methods, together with relevant data-driven optimisation and network analyses within this scope, and excluded studies not written in English, focused solely on conventional transport modes, or lacking clear quantitative methods or results relevant to these applications. Datasets recorded in other languages were included when they were used in English-language publications. \textcolor{black}{
Eligibility was determined by the research task, fitted model, and evaluation rather than by a strict distinction between statistical and \ac{ml} methods. Spatial statistical approaches, such as Gaussian processes, geographically weighted regression, and spatial autoregressive models, were not explicitly targeted by the original search terms. Their limited representation in the reviewed corpus therefore should not be interpreted as evidence that such methods are absent from micromobility research.}

\color{black}

\begin{figure}[ht]
    \vspace{-0.1in}
    \centering
    \includegraphics[width=\linewidth]{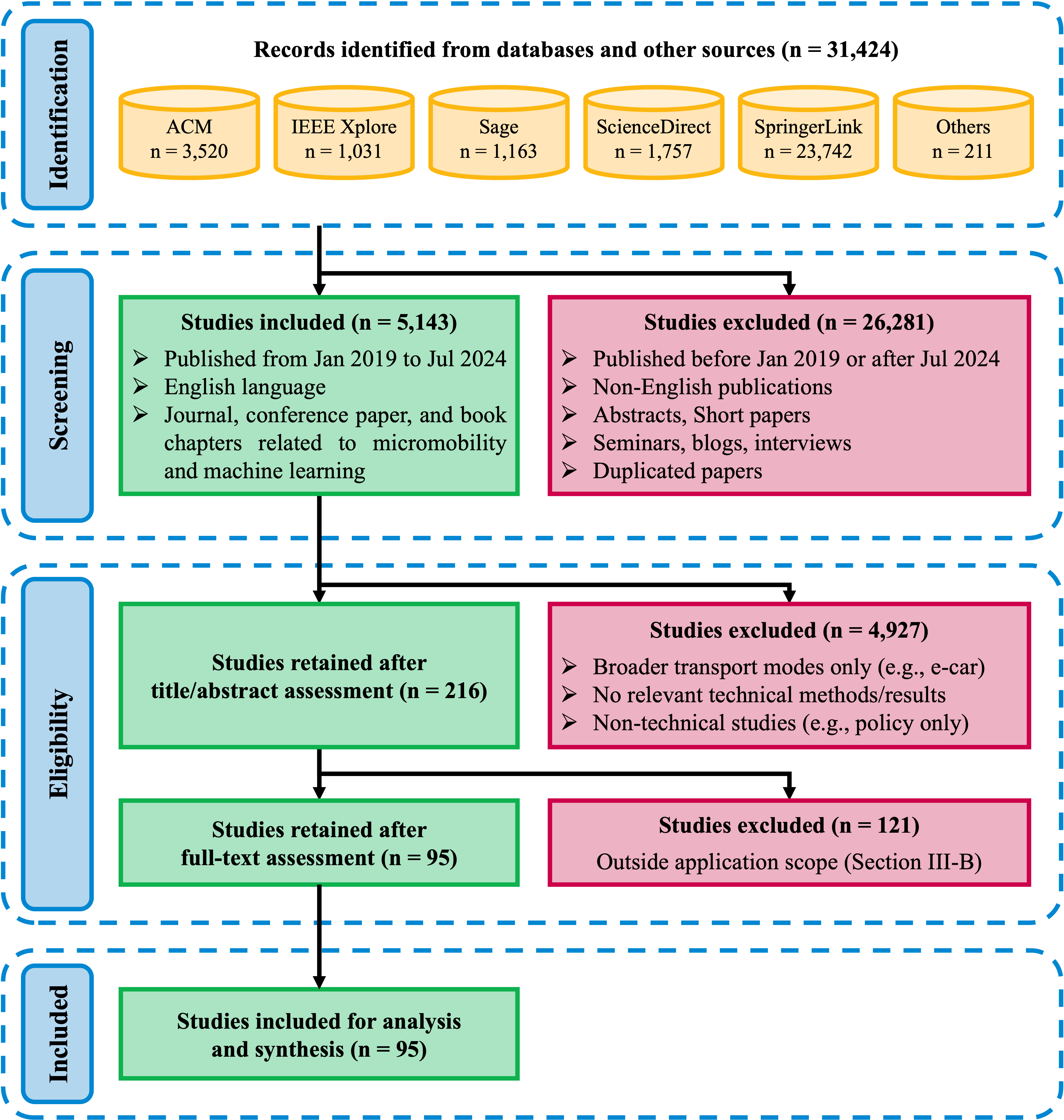}
    \caption{PRISMA flow diagram for literature review in our study.}
    \label{fig: prisma}
    \vspace{-0.1in}
\end{figure}

\subsection{Search Strategy}

We searched the following databases for literature related to computer science and data science: ACM Digital Library (Association for Computing Machinery), IEEE Xplore (Institute of Electrical and Electronics Engineers), MDPI (Multidisciplinary Digital Publishing Institute), SAGE Journals (SAGE Publishing), ScienceDirect (Elsevier), and SpringerLink (Springer Nature). Besides, we also used Google Scholar to capture relevant papers that may not be indexed in the databases above, and Kaggle was consulted for publicly available micromobility datasets.


\color{black}

To ensure comprehensive coverage, multiple synonymous terms and spelling variants were considered during the search process. The core Boolean search structure combined three groups of terms as follows:

\begin{quote}
    \textit{(``micromobility'' \textbf{OR} ``micro-mobility'' \textbf{OR} ``bike'' \textbf{OR} ``bicycle'' \textbf{OR} ``e-bike'' \textbf{OR} ``bike-sharing'' \textbf{OR} ``e-scooter'' \textbf{OR} ``e-mobility'') \\
    \textbf{AND} \\
    (``machine learning'' \textbf{OR} ``deep learning'' \textbf{OR} ``neural network'' \textbf{OR} ``computer vision'') \\
    \textbf{AND} \\
    (``dataset'' \textbf{OR} ``trip history'' \textbf{OR} ``demand'' OR ``resource-constrained'' \textbf{OR} ``energy consumption'' \textbf{OR} ``safety'')}
\end{quote}

The search was restricted by publication date only, from January 1, 2019, to July 31, 2024. No additional search-field restrictions were applied. The Boolean query was used across the selected databases, with minor syntax adaptations where required by the individual search interfaces. The retrieved records were subsequently screened through title and abstract assessment followed by full-text eligibility assessment.

Screening was conducted by two reviewers with responsibilities divided by topic. Image- and video-based studies and related computer vision applications were screened separately from the remaining studies. Records were not independently assessed by both reviewers, so no inter-reviewer agreement statistic or formal disagreement-resolution procedure was applied.

\color{black}


\autoref{fig: prisma} summarises the review process. In total, 31,424 records were identified. After initial screening, 5,143 studies met the criteria for publication date, language, publication type, and general relevance to micromobility and \ac{ml}. \textcolor{black}{
Title and abstract assessment reduced this set to 216 potentially relevant studies. Full-text assessment applied the application-focused scope in \autoref{sec: method}, retaining 95 studies and excluding the remaining 121 from the focused synthesis. A study-level evidence table for the 95 included studies, with full bibliographic references indexed S01–S95, is provided in the \textit{Supplementary Material}.}

\color{black}


\section{Datasets}
\label{sec: dataset}
\begin{figure}[ht]
    \vspace{-0.1in}
    \centering
    \includegraphics[width=\linewidth]{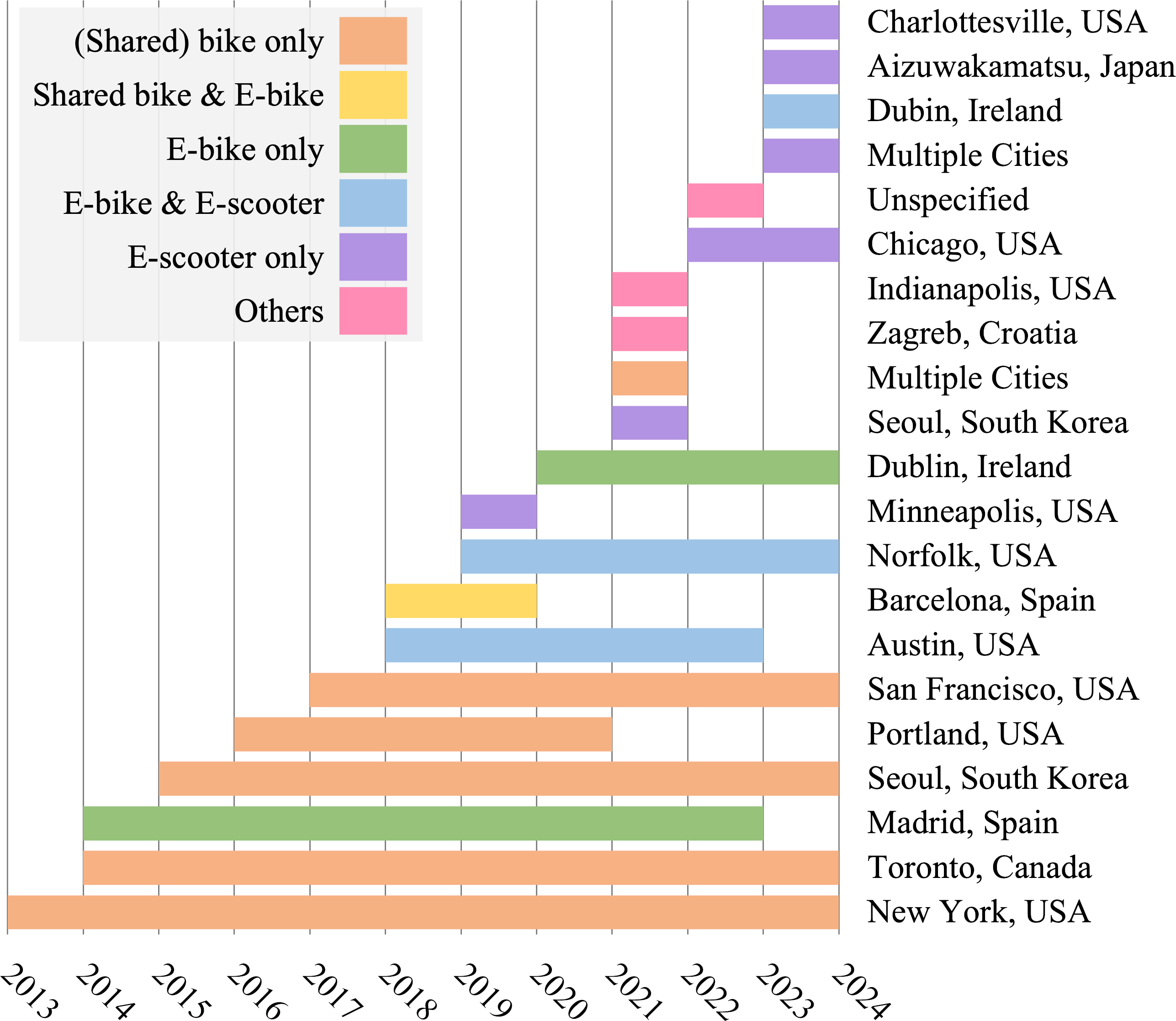}
    \caption{An overview of the micromobility datasets reviewed.}
    \label{fig: tabular dataset}
    \vspace{-0.2in}
\end{figure}

This section introduces the micromobility datasets identified in our review and summarises them in \autoref{table: datasets}, including \textcolor{black}{
15 tabular, 6 image/video, and 1 multimodal dataset} for different research tasks. An overview of these datasets is displayed in \autoref{fig: tabular dataset}. Generally speaking, the tabular datasets \cite{Lyft2024, Madrid2022, Toronto2017, Lyft2019, Seoul2018, Barcelona2018, Minneapolis2019, Norfolk2019, Austin2018, Biketown2018, Moby2020, CYCLANDS, Chicago2022, Erdeli2023, Ding2024} usually contain data from multiple sensors attached to micromobility or installed at parking spots, while the image datasets \cite{Nienaber2015_1, Nienaber2015_2, Apurv2021, Su2024, Chen2024, Kaundanya2024, Sabri2024} include pictures or video frames from cameras attached to micromobility or other vehicles. They are sourced from various micromobility devices around the world.

\color{black}


To provide a structured understanding of these datasets, we analyse them from multiple complementary perspectives: \textbf{(1) Spatial \& Temporal Analysis}, \textbf{(2) Vehicle Type Analysis}, \textbf{(3) Tabular Dataset Feature Analysis}, and \textbf{(4) Image-Based Dataset Analysis}. These perspectives are intended to highlight how different dataset properties support specific analytical objectives and machine learning applications in micromobility research.

\color{black}

\subsection{Spatial \& Temporal Analysis}


\color{black}

\textcolor{black}{
Among the 21 datasets for which temporal coverage is shown in \autoref{fig: tabular dataset}, four of the longest-running datasets are} \cite{Lyft2024, Madrid2022, Toronto2017, Seoul2018}, \textcolor{black}{with several remaining ongoing.} These datasets are mainly provided by micromobility operators through local open-data portals and provide valuable long-term records for analysing temporal trends and system changes. \textcolor{black}{Twelve datasets span two years or less, with the duration of ongoing datasets calculated up to July 31, 2024.} These shorter datasets are more commonly associated with specific research objectives or recent methodological studies, reflecting growing research interest in micromobility.


From a spatial perspective, the reviewed datasets cover North America, Europe, and Asia. The USA contributes datasets from several cities, including New York, San Francisco, Austin, and Chicago. Smaller cities are also represented \cite{Erdeli2023, Su2024, Chen2024}, showing that micromobility data collection is not limited to major metropolitan areas.


Some datasets do not specify an exact collection location. The Pothole dataset \cite{Nienaber2015_1, Nienaber2015_2}, for example, was compiled at Stellenbosch University, but its collection location was not reported and is therefore excluded from \autoref{fig: tabular dataset}. Two datasets are labelled ``Multiple Cities''. CYCLANDS \cite{CYCLANDS} and the Micromobility Lane Recognition Dataset \cite{Kaundanya2024} cover multiple geographical areas, while PolyMMV \cite{Sabri2024} was constructed from publicly available videos with unspecified geographical coverage. \textcolor{black}{
However, geographic coverage across multiple locations does not itself indicate that a model has been trained and evaluated for cross-city or cross-operator generalisation.}

\color{black}

\subsection{Vehicle Type Analysis}


\color{black}

From the perspective of vehicle type, 6 of the 21 datasets reviewed focus on bicycles or shared bikes, including long-running datasets from Toronto and Seoul. \textcolor{black}{
These datasets provide relatively long-term records of bicycle and shared-bike use. Similarly, e-scooter datasets are also well represented in the reviewed literature} \cite{Norfolk2019, Minneapolis2019, Chicago2022, Su2024, Chen2024}. Several datasets cover multiple vehicle types \cite{Barcelona2018, Austin2018, Norfolk2019, Ding2024, Sabri2024}, including combinations of bicycles, e-bikes, and e-scooters. \textcolor{black}{
These mixed-mode datasets enable comparisons across different micromobility vehicle types.}


Three datasets are labelled ``Others'' in \autoref{fig: tabular dataset}. The dataset in \cite{Erdeli2023} was collected through a smartphone application and covers several travel modes, including walking, cycling, and driving. The dataset in \cite{Apurv2021} contains images of pedestrians and e-scooter riders captured from cars, while \cite{Sabri2024} integrates publicly available videos rather than using a dedicated vehicle-mounted collection platform.

\color{black}

\subsection{Tabular Dataset Feature Analysis}

\color{black}


Most tabular datasets consist of trip or usage records containing four main feature groups: user and vehicle information (e.g., user ID and vehicle type), spatial and temporal data (e.g., departure time and arrival position), trip attributes (e.g., distance and duration), and operational variables (e.g., battery state of charge).

The user and vehicle information \textcolor{black}{recorded varies across data providers} but can support analysis of usage patterns and user behaviour. Spatial and temporal features capture mobility dynamics such as peak usage periods, seasonal variation, and spatial demand patterns, while trip attributes describe journey characteristics and system use. Operational variables support fleet and energy management. Among the \textcolor{black}{
reviewed electric micromobility datasets}, the Moby Bikes dataset \cite{Moby2020} records battery state of charge before and after trips, while \cite{Ding2024} provides real-time battery state-of-charge information.


\color{black}

\subsection{Image-Based Dataset Analysis}

\color{black}


Image-based micromobility datasets are mainly developed for safety-related perception tasks using cameras mounted on e-scooters, e-bikes, or other vehicles. Datasets such as \cite{Apurv2021, Sabri2024} support object detection of pedestrians, vehicles, traffic lights, signs, and other micromobility users. Other datasets include road and infrastructure features such as lanes, potholes, bumps, fences, walls, and traffic cones \cite{Su2024, Kaundanya2024, Nienaber2015_1, Nienaber2015_2}. The Micromobility Lane Recognition dataset \cite{Kaundanya2024}, for example, is designed for multilabel classification of the riding environment to support real-time lane recognition.


Existing open-source autonomous-driving datasets \cite{Geiger2013, Cheng2019, Chang2019, Xiao2021, Sun2020, Caesar2020} can provide relevant visual information but are not directly transferable to micromobility because of differences in sensing perspective, vehicle motion, and operating environment. E-scooters experience frequent vibration and operate under diverse lighting, weather, and road conditions, which can further affect perception performance.


Nevertheless, these datasets \textcolor{black}{
remain useful resources for safety-related research, although autonomous-driving datasets require micromobility-specific validation before direct application to micromobility scenarios}.

\color{black}

\begin{table*}[ht]
    \caption{Datasets.}
    \vspace{-0.1in}
    \label{table: datasets}
    \scriptsize
    \begin{threeparttable}
    \begin{tabularx}{\linewidth}{@{\extracolsep{\fill}}X c c c c c c c}
        \toprule
        \textbf{Dataset Name} & \textbf{Year} & \textbf{Size} & \textbf{Vehicle Type} & \textbf{Purpose} & \textbf{Data Modalities} & \textbf{Labels or Annotations} & \textbf{Availability} \\
        
        \midrule
        \multicolumn{8}{c}{\cellcolor[RGB]{204, 239, 252}\textbf{Structured Tabular Datasets}} \\
        
        \midrule
        \makecell[X]{New York Lyft Citi Bike Trip Data \cite{Lyft2024}} & 2013 & \makecell{700 MB \\per month} & shared bike & \makecell{demand \\prediction} & trip record & \makecell[c]{``\textit{Start Station ID}'', \\``\textit{End Station ID}''} & public \\
        \midrule
        \makecell[X]{Madrid BiciMAD E-bike Daily Usage Record\textsuperscript{*} \cite{Madrid2022}} & 2014 & 4.35 MB & e-bike & general & usage record & ``\textit{Total Uses}'' & public \\
        \midrule
        \makecell[X]{Bike Share Toronto Ridership Data \cite{Toronto2017}} & 2014 & \makecell{60 MB\\per month} & shared bike & \makecell{demand \\prediction} & trip record & \makecell[c]{``\textit{Trip Start Station ID}'', \\``\textit{Trip End Station ID}''} & public \\
        \midrule
        \makecell[X]{San Francisco Lyft Bay Wheels E-Bike Dataset \cite{Lyft2019}} & 2017 & \makecell{6 MB \\per month} & shared bike & general & trip record & \makecell[c]{``\textit{Start Station ID}'', \\``\textit{End Station ID}''} & public \\
        \midrule
        \makecell[X]{Seoul Bike Sharing Rental History\textsuperscript{*} \cite{Seoul2018}} & 2017 & \makecell{600 MB \\per season} & shared bike & \makecell{demand \\prediction} & rental record & \makecell[c]{``\textit{Rental Station ID}'', \\``\textit{Return Station ID}''} & public \\
        \midrule
        \makecell[X]{Barcelona Bicing Shared Bike Usage Record \cite{Barcelona2018}} & 2018 & 1.57 MB & \makecell{shared bike \\e-bike} & general & usage record & \makecell[c]{``\textit{bikesInUsage}'', \\``\textit{electricalBikesInUsage}''} & public \\
        \midrule
        \makecell[X]{Motorized Foot Scooter Trips Dataset \cite{Minneapolis2019}} & 2019 & \makecell{15.5 MB \\per month} & e-scooter & general & trip record & \makecell[c]{``\textit{StartCenterlineID}'', \\``\textit{EndCenterlineID}''} & public \\
        \midrule
        \makecell[X]{Norfolk Lime Micromobility Dataset \cite{Norfolk2019}} & 2019 & 300 MB & \makecell{e-scooter, \\e-bike} & general & trip record & \makecell[c]{``\textit{Starting Tract}'', \\``\textit{Ending Tract}''} & public \\
        \midrule
        \makecell[X]{Austin Shared Mobility Vehicle Trips \cite{Austin2018}} & 2019 & 3.7 GB & \makecell{e-scooter, \\e-bike} & \makecell{demand \\management} & trip record & \makecell[c]{``\textit{Census Tract Start}'', \\``\textit{Census Tract End}''} & public \\
        \midrule
        \makecell[X]{Portland Biketown Trip Data \cite{Biketown2018}} & 2020 & \makecell{6 MB\\ per month} & shared bike & \makecell{demand \\prediction} & trip record & ``\textit{StartHub}'', ``\textit{EndHub}'' & public \\
        \midrule
        \makecell[X]{Moby Bikes Bikeshare Data \cite{Moby2020}} & 2020 & \makecell{10 MB\\ per month} & e-bike & general & GPS record & \makecell[c]{``\textit{LastGPSTime}'', \\``\textit{Latitude}'', ``\textit{Longitude}''} & public \\
        \midrule
        \makecell[X]{CYCLANDS: CYCling geo-Located AccideNts, their Details and Severities \cite{CYCLANDS}} & 2021 & \makecell{18 kB to\\ 29 MB} & bicycle & rider safety & \makecell{accident \\record} & \makecell[c]{``\textit{Location Details}'', \\``\textit{Personal Characteristics}'', \\``\textit{Vehicle Characteristics}''}& public \\
        \midrule
        \makecell[X]{Chicago E-scooter Trip Data \cite{Chicago2022}} & 2022 & 1.1 GB & e-scooter & general & trip record & \makecell[c]{``\textit{Start Centroid Location}'', \\``\textit{End Centroid Location}''}& public \\
        \midrule
        \makecell[X]{Collecty Dataset \cite{Erdeli2023}} & 2023 & 24.86 GB & \makecell{multimodal \\transport} & \makecell{mobility \\analysis} & \makecell{activity \\details} & mobility analysis & public \\
        \midrule
        \makecell[X]{Dublin Electric Micromobility Dataset \cite{Ding2024}} & 2024 &  7.2 MB & \makecell{e-scooter \\e-bike} & \makecell{energy \\management} & trip record & ``\textit{SoC}'' & public \\
        
        \midrule
        \multicolumn{8}{c}{\cellcolor[RGB]{204, 239, 252}\textbf{Image \& Video Datasets}} \\
        
        \midrule
        \makecell[X]{Road Pothole Images for Pothole Detection \cite{Nienaber2015_1, Nienaber2015_2}} & 2015 & \makecell{9.5 GB} & car & \makecell{rider safety} & image & potholes & public \\
        \midrule
        \makecell[X]{IUPUI‑CSRC E‑Scooter Rider Detection Benchmark Dataset \cite{Apurv2021}} & 2021 & \makecell{536 MB} & car & \makecell{rider safety} & image & e-scooter riders, pedestrians & public \\
        \midrule
        \makecell[X]{Outdoor Hazard Detection Dataset \cite{Su2024}} & 2023 & \makecell{13.27 GB} & e-scooter & \makecell{rider safety} & image & outdoor hazard & public \\
        \midrule
        \makecell[X]{ScooterDet Dataset \cite{Chen2024}} & 2024 & \makecell{1.1 GB} & e-scooter & \makecell{rider safety} & image & road objects & public \\
        \midrule
        \makecell[X]{Micromobility Lane Recognition Dataset \cite{Kaundanya2024}} & 2024 & \makecell{18.10 GB} & e-scooter & \makecell{micromobility \\ safety} & image & lanes & public \\
        \midrule
        \makecell[X]{PolyMMV Dataset \cite{Sabri2024}} & 2024 & 298 MB & \makecell{e-scooter, \\e-bike} & \makecell{rider safety} & video & \makecell{micromobility \\vehicles} & public \\
        
        \midrule
        \multicolumn{8}{c}{\cellcolor[RGB]{204, 239, 252}\textbf{Multimodal Datasets}} \\
        
        \midrule
        \makecell[X]{Real-time Vibration Sensor Dataset for E-scooter \cite{Kim2022}} & 2021 & \makecell{NA} & e-scooter & \makecell{rider safety} & multimodal & \makecell{riding behaviour} & private \\
        \bottomrule
    \end{tabularx}
    \begin{tablenotes}\footnotesize
        \item[1] For image datasets, the ``Label or Annotations'' column shows the target object(s) in the image, while for tabular datasets, it shows the key column name(s).
        \item[2] Datasets marked with \textbf{*} were recorded in other languages and used in English publications. For the convenience of readers, the column names are translated into English and presented in the corresponding ``Label or Annotations'' column.
    \end{tablenotes}
    \end{threeparttable}
    \vspace{-0.2in}
\end{table*}


\section{Techniques}
\label{sec: tech}


\textcolor{black}{This section summarises the \ac{ml} methods identified in the reviewed papers, as shown in \autoref{table: techniques}, ranging from single-model approaches (e.g., \ac{lr} and \ac{gcn}) to hybrid and multi-stage frameworks. The categorisation is literature-driven and reflects how these methods are used and reported in micromobility studies rather than a comprehensive taxonomy of \ac{ml}.}

\begin{table*}[ht]
    \caption{Techniques.}
    \vspace{-0.1in}
    \label{table: techniques}
    \scriptsize
    \begin{adjustbox}{width=\textwidth}
    \begin{tabularx}{\linewidth}{@{\extracolsep{\fill}}l c c l X c}
        \toprule
        \textbf{Technique Family} & \textbf{Representative Methods} & \textbf{Typical Modality} & \textbf{Primary Use Cases} & \textbf{Strengths \& Limitations} & \textbf{References}\\
        \midrule
        Statistics-based & LR, MLR, ARIMA & structured features & \makecell[l]{demand prediction \\energy consumption prediction} & \makecell[l]{Strong interpretability and simple baselines \\Sensitive to outliers; limited nonlinearity} & \cite{Saum2020, Gioldasis2024} \\
        \midrule
        Distance-based & kNN, SVM, K-means & structured features & \makecell[l]{driving behaviour clustering \\dispatching clustering} & \makecell[l]{Few distributional assumptions; kernels capture \\nonlinearity \\Sensitive to metric/kernel; costly at scale} & \cite{Masoud2023, Ina2022} \\
        \midrule
        Tree-based & DT, RF, XGB, LGBM & structured features & \makecell[l]{demand prediction \\energy consumption prediction} & \makecell[l]{Captures nonlinearity and feature interactions; partly \\interpretable \\Higher complexity and tuning cost} & \cite{Ullah2021_1, Ouf2023} \\
        \midrule
        MLP & MLP & structured features & \makecell[l]{destination prediction \\energy consumption prediction} & \makecell[l]{Simple and effective for tabular data \\Weak inductive bias for complex data structures} & \cite{Liang2022, Abdelaty2021} \\
        \midrule
        CNN-based & CNN, MobileNetV2 & images, videos & \makecell[l]{lane recognition \\object detection} & \makecell[l]{Strong local feature extraction; supports lightweight \\edge models \\Often label-hungry; compute-accuracy trade-off} & \cite{Kaundanya2024, Kim2022} \\
        \midrule
        RNN-based & RNN, LSTM, GRU & time series & \makecell[l]{trajectory prediction \\demand prediction} & \makecell[l]{Models temporal dynamics effectively \\Costly training and latency for long sequences} & \cite{Liang2022, Nawaz2020} \\
        \midrule
        GNN-based & GCN, STGCN & graphs, networks & \makecell[l]{demand prediction \\traffic flow prediction} & \makecell[l]{Uses topology and neighbourhood dependencies\\Relies on graph structure; high deployment cost} & \cite{Xiao2020, Miao2021} \\
        \midrule
        Transformers & \makecell{Vision Transformer, \\Transformer encoders} & images, time series & \makecell[l]{demand prediction \\lane detection} & \makecell[l]{Self-attention captures long dependencies; flexible \\architectures \\High memory and compute cost on edge devices} & \cite{Xu2023, Dosovitskiy2020} \\
        \midrule
        \makecell[l]{Hybrid \& multi- \\stage frameworks} & \makecell{STFT + CNN, \\optical flow + detector, \\segmentation + detector} & \makecell{multimodal sensors \\(vibration+vision)} & \makecell[l]{multimodal fusion \\multi-task safety pipelines \\flow + detection \\segmentation + tracking} & \makecell[l]{Combines complementary signals for robust real-time \\decisions \\Pipeline complexity and error propagation risks} & \cite{Su2024, Subramanyam2023} \\
        \bottomrule
    \end{tabularx}
    \end{adjustbox}
    \vspace{-0.2in}
\end{table*}


\subsection{\ac{ml} Process in Micromobility}

\begin{figure}[ht]
    \vspace{-0.1in}
    \centering
    \includegraphics[width=\linewidth]{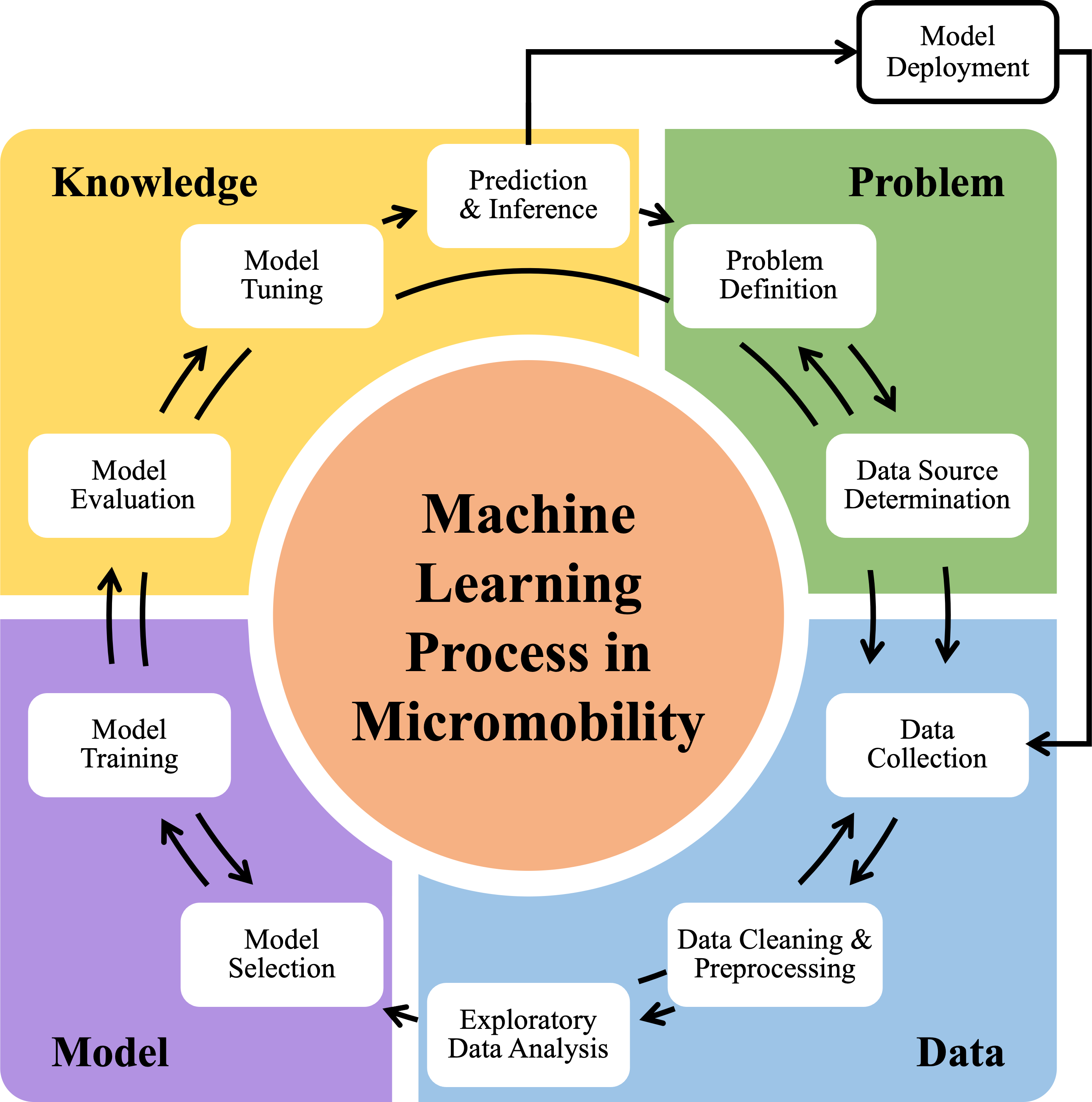}
    \caption{A typical \ac{ml} workflow as observed in micromobility studies.}
    \label{fig: ml process}
    \vspace{-0.1in}
\end{figure}

This subsection provides an overview of the typical \ac{ml} workflow as observed in the reviewed micromobility literature. \textcolor{black}{As shown in \autoref{fig: ml process}, the workflow is organised into four conceptual dimensions, i.e., problem, data, model, and knowledge, following prior studies \cite{He2021, Amershi2019}. It covers problem formulation, data preparation, model development, evaluation, and deployment, while reflecting the practical requirements and data characteristics of micromobility systems.}

\color{black}

The workflow begins with \textbf{Problem Definition}, where the target variable and objectives of the \ac{ml} task are specified according to the micromobility application. This is followed by \textbf{Data Source Determination} and \textbf{Data Collection}, during which relevant sources, such as vehicle-to-everything communication systems, roadside sensors, and weather monitoring stations, are identified, assessed, extracted, and organised. During \textbf{Data Preprocessing \& Cleaning}, missing values, outliers, duplicated records, GPS errors, and temporal or spatial inconsistencies are addressed, together with transformations and feature engineering where required. \textbf{Exploratory Data Analysis} then examines data distributions and relationships to identify spatial, temporal, and behavioural patterns and to inform subsequent model development.

Based on the prepared data, \textbf{Model Selection} considers the task, data availability, spatial and temporal resolution, and deployment requirements. During \textbf{Model Training}, model parameters are learned from the available training data, followed by \textbf{Model Evaluation} using unseen data and task-specific metrics. Classification tasks commonly use accuracy, precision, recall, and F1 score, while regression tasks report measures, e.g., mean absolute error, root mean square error, and $R^2$. For micromobility applications, evaluation should also consider human and operational factors, including asymmetric error costs, e.g., missed hazards versus false alarms, and whether outputs are sufficiently timely, stable, and interpretable to support decision-making. \textbf{Model Tuning} further refines hyperparameters, features, or model configurations, after which \textbf{Prediction \& Inference} applies the resulting model to unseen data to support the original application objective.

Finally, \textbf{Model Deployment} introduces additional challenges because micromobility systems are affected by policy changes, infrastructure updates, weather and seasonality, fleet strategies, and shifting user demographics \cite{Dozza2022}. Deployed models may require ongoing monitoring and periodic updates, e.g., through retraining or rolling-window approaches \cite{Lyu2024}. For streaming settings, online or continual learning may also be considered together with time-aware evaluation to maintain valid performance assessment as operating conditions change.

\color{black}

\subsection{Classical \ac{ml} Models}

In this study, we divided classical \ac{ml} models included in this paper based on their core principles and primary objectives into statistics-based, distance-based, and tree-based models. Models in each category have their own characteristics and are used in different situations.




\subsubsection{Statistics-Based Models}

Statistical models, such as \ac{lr}, \ac{mlr}, and \ac{arima}, are based on statistical theory and use probability models to describe and predict targets. \textcolor{black}{
Their assumptions concern the relation between predictors,
targets, and errors, with parameter estimation or hypothesis
testing depending on the task \cite{Montgomery2021, Jani2023, Li2023}}. \textcolor{black}{For example, \ac{lr} was used as a baseline for battery range prediction \cite{Li2024}, \ac{mlr} for e-mobility energy consumption prediction \cite{Ullah2021}, and \ac{arima} and seasonal \ac{arima} for destination and hourly demand prediction \cite{Miao2021, Saum2020}. While linear specifications restrict functional form, statistical models can also represent nonlinear or spatial structures}.


\textcolor{black}{
Spatial statistical approaches, such as Gaussian processes, geographically weighted regression, and spatial autoregressive models, were not explicitly targeted by the original search terms and are therefore not comprehensively represented in the reviewed corpus.}

\subsubsection{Distance-Based Models}


\textcolor{black}{Distance-based models rely on proximity or similarity between data points for classification or clustering. Methods such as \ac{knn} and K-Means operate using explicit distance metrics, such as Euclidean or Manhattan distance, while \ac{svm} can represent similarity through kernel functions in a transformed feature space. The choice of distance metric or kernel therefore affects neighbourhood structure, cluster assignments, or decision boundaries, particularly for nonlinear or high-dimensional data \cite{Moumen2020, Faizi2024}. In micromobility studies, \ac{knn} was used as a baseline for trip destination prediction \cite{Liang2022}, K-Means for e-scooter vehicle routing \cite{Masoud2023}, and \ac{svm} for e-scooter battery range prediction \cite{Li2024}. However, these methods can be computationally intensive and sensitive to the selected distance metric or kernel, while noisy or poorly separated data may further reduce their performance.}

\subsubsection{Tree-Based Models}


\color{black}

\ac{dt} and tree-based ensemble methods, including \ac{rf}, \ac{xgb}, and \ac{lgbm}, model nonlinear relationships and feature interactions through recursive partitioning of the feature space \cite{Liu2019_1, Bittencourt2024}. In micromobility studies, \ac{dt} and \ac{rf} were used as baseline models for e-mobility energy consumption prediction \cite{Ullah2021_1}, while \ac{xgb} and \ac{lgbm} were applied to e-bike energy needs estimation \cite{Ouf2023}. Tree-based ensemble models can require greater computational resources than simpler baseline models, particularly as the number and depth of trees increase.

\subsection{Artificial Neural Networks}

Based on their architectures and the types of data they process, we discuss \ac{mlp}, \ac{cnn}, \ac{rnn}, \ac{gnn}, and Transformer models commonly used in micromobility research.


\subsubsection{Multi-Layer Perceptron}


\ac{mlp} consists of multiple fully connected layers and is particularly suitable for structured or tabular data, but lacks specialised mechanisms for modelling spatial or temporal dependencies \cite{Borisov2024, Bratsas2019}. In micromobility studies, \ac{mlp} were used as a baseline for destination prediction \cite{Liang2022} and for modelling e-mobility energy consumption \cite{Abdelaty2021}.

\subsubsection{Convolutional Neural Networks} 


They use convolutional operations to extract local spatial features, making it well suited to image and video data. In micromobility research, \ac{cnn} models have been applied to safety-related tasks including real-time lane recognition \cite{Kaundanya2024, Jang2023}, rider attentiveness prediction \cite{Kim2022}, rider fall detection \cite{Nguyen2021}, object detection \cite{Nienaber2015_1, Nienaber2015_2}, and object tracking and trajectory estimation \cite{Yamaguchi2022}. Beyond native visual inputs, signals can also be transformed into 2D representations, such as spectrograms or time-frequency maps, before being analysed using \ac{cnn} architectures \cite{Lee2022, Jiang2019, Modi2020, Yi2021}. This enables local temporal--frequency patterns to be captured using established image-based network designs.

\subsubsection{Recurrent Neural Networks}


\ac{rnn} models, including Simple \ac{rnn}, \ac{lstm}, \ac{gru}, and their variants, are designed to model sequential and temporal dependencies \cite{Sehovac2020, Hewamalage2021}. Simple \ac{rnn} can struggle with long-term dependencies, while \ac{lstm} and \ac{gru} use gating mechanisms to improve information retention, with \ac{gru} providing a simpler architecture than \ac{lstm}. Convolutional \ac{lstm} further incorporates spatial processing, making it suitable for spatiotemporal data at the cost of additional computational complexity. In the reviewed literature, Simple \ac{rnn} and \ac{lstm} were used in \cite{Liang2022}, while \ac{gru} and Convolutional \ac{lstm} were applied in \cite{Nawaz2020} to predict trip destinations from spatiotemporal sequences.

\subsubsection{Graph Neural Networks}


\ac{gnn} models, including \ac{gcn}, \ac{stgcn}, and other variants, are designed to process graph-structured data \cite{Chen2021, Wu2022}. \ac{gcn} aggregates information from neighbouring nodes through graph convolution, allowing relationships between connected nodes to be modelled, although deeper architectures may suffer from over-smoothing as node representations become increasingly similar. \ac{stgcn} extends this framework by incorporating temporal dependencies, making it particularly suitable for spatiotemporal tasks such as traffic flow and mobility demand prediction. \textcolor{black}{
An important distinction among graph-based models lies in how relationships between locations are represented. Earlier multi-graph approaches to bike-sharing demand prediction constructed multiple graphs to capture heterogeneous inter-station relationships \cite{Chai2018}, while adaptive graph methods such as Graph WaveNet learn latent dependencies directly from data rather than relying solely on a predefined graph structure \cite{Wu2019}. Within the reviewed micromobility literature, \ac{gcn} has been employed for trip destination prediction in \cite{Miao2021}, while \ac{stgcn} was used to predict hourly shared-bike demand in \cite{Xiao2020}. These studies illustrate that graph construction itself is an important modelling choice, particularly when spatial proximity alone does not fully represent demand interactions.}

\subsubsection{Transformers}

Transformer models use self-attention to capture long-range dependencies. Transformer-based sequence models have been explored for demand forecasting \cite{Xu2023, Xu2023_2, Rhmann2024} and battery range estimation \cite{Li2024}. In \cite{Xu2023}, the Transformer-based model achieved the lowest mean absolute error for pickups and returns, but a graph recurrent baseline achieved lower root mean square error. Its reported superiority therefore depended on the evaluation metric. \textcolor{black}{
The multi-graph Transformer in \cite{Xu2023_2} jointly represents heterogeneous spatial relationships for dockless scooter demand forecasting in the USA. However, the evaluations use within-city temporal splits, so results from the two cities do not demonstrate cross-city or cross-operator generalisation. Similarly, a temporal fusion Transformer is applied to bike-sharing activity prediction in \cite{Rhmann2024}, but its reported advantage over recurrent models remains specific to the selected aggregation interval, forecast horizon, dataset, and evaluation protocol.} Vision Transformers extend the architecture to image analysis by representing images as sequences of patches \cite{Dosovitskiy2020}. Related studies have applied Transformer-based vision models to lane detection and road scene understanding \cite{Peng2023, Chen2022, Liu2021, Bai2023}, \textcolor{black}{
providing relevant methodological background for micromobility perception tasks.} Computational and memory costs can constrain deployment on micromobility devices \cite{Liu2024}.

\color{black}


\subsection{Hybrid \& Multi-Stage \ac{ml} Frameworks}

\color{black}

Hybrid and multi-stage \ac{ml} frameworks combine multiple modelling components in sequence or parallel, allowing the output of one stage to support subsequent tasks. In micromobility research, such frameworks are commonly used for multimodal data and multi-task applications, e.g., vibration analysis, motion estimation, segmentation, and object detection under real-time or resource-constrained conditions.


\subsubsection{Data Transformations \& Image Processing}


Data transformations can convert non-image signals into representations suitable for image-based models. In \cite{Kim2022}, short-time Fourier and wavelet transforms were used to convert e-scooter vibration signals into images, which were then analysed using multimodal \ac{cnn} models for rider-attentiveness monitoring. \textcolor{black}{
This study illustrates how complementary representations of vibration signals can be combined within a multimodal learning framework for rider-attentiveness monitoring.}

\color{black}

\subsubsection{Optical Flow \& Object Detection}

Optical flow is a technique used to describe the apparent position of objects of interest by tracking reference points across multiple frames over a certain period and calculating the displacement of corner points among these frames. Combined with object detection techniques, optical flow can approximate the speed of e-scooters and predict potential collision times with stationary objects. This idea has been employed to develop an accident prevention system in \cite{Subramanyam2023, Sabri2024} to provide real-time warnings to micromobility riders in situations of potential accidents.

\subsubsection{Semantic Segmentation \& Object Detection}

Combining semantic segmentation with object detection provides a balanced approach that enhances both accuracy and efficiency, especially in resource-limited scenarios. In \cite{Su2024}, these techniques were integrated using ``cells of interest'' to classify objects rather than traditional pixel-based methods. This approach reduces the computational load while maintaining high precision, making it suitable for micromobility applications, where accuracy and computational efficiency are both essential.


\section{Applications}
\label{sec: app}


\color{black}

\begin{figure}[ht]
    \vspace{-0.1in}
    \centering
    \includegraphics[width=\linewidth]{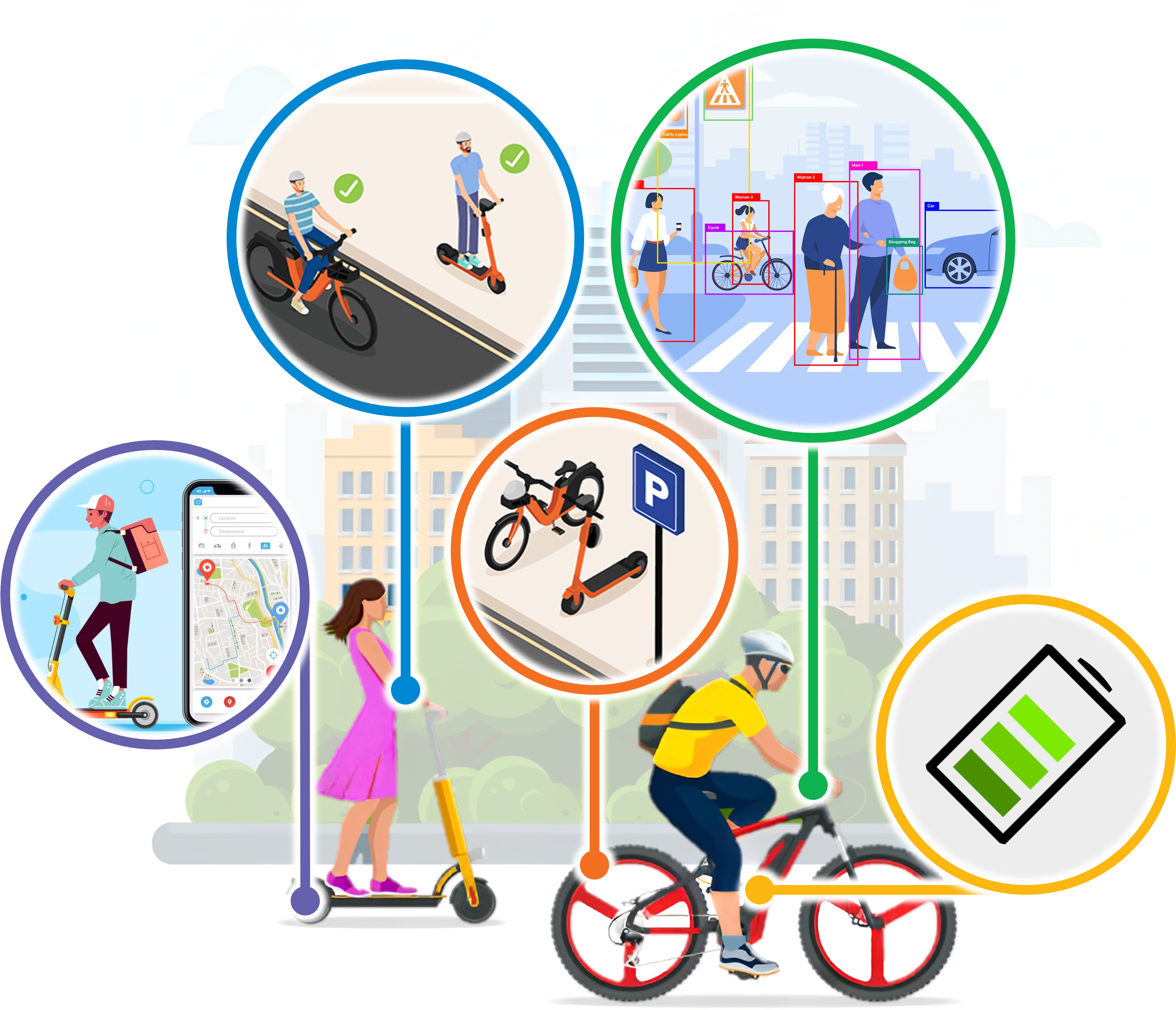}
    \caption{A pictorial representation of selected \ac{ml} applications in micromobility, such as safety and energy management, to improve efficiency, accuracy, and user experience.}
    \label{fig: ml app}
    \vspace{-0.2in}
\end{figure}

\textcolor{black}{
\ac{ml} methods have been applied to a range of prediction, optimisation, and safety-related tasks in micromobility systems.} These applications are summarised in \autoref{table: applications}, with selected examples illustrated in \autoref{fig: ml app}. They are grouped into \textit{Demand Prediction}, \textit{Trip Destination Prediction}, \textit{Energy Consumption Prediction}, \textit{Route Planning with Constraints}, and \textit{Safety}. Safety-related applications include \textit{Lane Recognition}, \textit{Object Detection}, \textit{Object Tracking}, and \textit{Trajectory Estimation}, which support collision-risk reduction, situational awareness, and real-time warning or intervention.

\color{black}

\subsection{Demand Prediction}


\textcolor{black}{Demand prediction is one of the most common \ac{ml} applications in micromobility, involving the forecasting of trip demand, vehicle usage, or availability across spatial and temporal intervals. It typically combines temporal forecasting with spatial analysis, while urban dynamics and regional scalability remain important challenges \cite{Yan2023, Abouelela2023}.} \textcolor{black}{
Existing studies have applied time-series models, neural networks, and graph-based methods \cite{Guidon2019, Zhou2022, Saum2020, Phithakkitnukooon2021, Boonjubut2022, Lee2022, Luo2021, Xiao2020, Saum2024, PelezRodrguez2024, Ma2020, Hua2020, Cuong2024, Yang2020, E2020, Shi2023}. Their performance depends on data completeness, spatial and temporal resolution, contextual variables, and implementation conditions \cite{Teusch2023, Albuquerque2021, Abouelela2023, Golpayegani2022}.}

\textcolor{black}{
Forecasting errors should be interpreted with the forecast horizon and spatial aggregation because these settings support different operational decisions. 
Observed trips also omit unmet demand when vehicles are unavailable, so evaluation should consider vehicle availability, unmet requests, rebalancing distance, and operating cost alongside prediction errors. Demand forecasting has been linked with rebalancing optimisation \cite{Saum2024}, providing decision-level evidence beyond forecast error, although not from a live operational trial.}


\subsection{Trip Destination Prediction}


\textcolor{black}{Trip destination prediction requires modelling temporal patterns, spatial relationships, and user-specific behaviours.} \textcolor{black}{
Time-series, tree-based, and neural network methods have been applied to this task \cite{Zong2019, Wang2021, Liu2019, Li2021, Chang2020, Tsiligkaridis2022, Zhao2022, Zhang2023_1}. However, the reported performance depends on the prediction target, spatial representation, available user information, and evaluation protocol, so results across different destination-prediction settings are not directly comparable. 
Detailed trajectories and user-specific mobility histories also raise privacy concerns because they can reveal individual travel patterns \cite{Himthani2020, Yang2024}.}

\subsection{Energy Consumption Prediction}


\textcolor{black}{
\ac{ml} methods have been applied to energy consumption prediction in electric micromobility vehicles.} \textcolor{black}{This task is typically formulated as regression or time-series forecasting using driving conditions, weather, battery characteristics, and trip attributes.} \textcolor{black}{
Linear models, ensemble methods, and neural networks have been investigated \cite{Ouf2023, Ina2022, Gioldasis2024}, with predictions supporting battery management and route planning \cite{Pokharel2021, Ullah2021, Nabi2023}. However, differences in vehicle type, battery capacity, rider behaviour, route profile, weather, and target definition limit direct comparison across studies and mean that prediction accuracy alone does not demonstrate operational or energy-saving benefits \cite{Nabi2023, Yan2023_1}.}

\subsection{Route Planning with Constraints}


\textcolor{black}{Route planning has been investigated under constraints including travel time \cite{Aboeleneen2024}, air quality \cite{Yan2023_2}, and trip mode preferences \cite{Shah2024}.} \textcolor{black}{
Electric micromobility may additionally require consideration of battery state and remaining range.} \textcolor{black}{These problems typically combine prediction of quantities such as travel time or energy consumption with optimisation of competing objectives and constraints.} \textcolor{black}{
Statistics-based, tree-based, and neural network models have been investigated as predictive components \cite{Scott2021, Masoud2023, Li2024}. Route quality also depends on how dynamic constraints and competing objectives are represented \cite{Lv2022, Hou2023}. Lower prediction error therefore does not necessarily imply a better route without evaluation against the intended operational objectives.}

\subsection{Lane Recognition}


\textcolor{black}{Lane recognition identifies whether a micromobility rider is travelling in a designated lane or an unauthorised area. This is particularly relevant because e-scooter riders may use sidewalks due to traffic congestion, poor infrastructure, or simply for amusement, increasing collision risk with pedestrians.} \textcolor{black}{
Lightweight models such as MobileNetV2 have been investigated for lane recognition on resource-constrained devices \cite{Kaundanya2024, Sandler2018}. However, classification performance and model size alone do not establish deployability. Target-device latency, memory and energy requirements, and end-to-end warning performance also require evaluation.}

\subsection{Object Detection}


\textcolor{black}{Object detection supports hazard recognition by localising and classifying moving and stationary objects in images or videos.} \textcolor{black}{
The purpose is to provide information about moving and stationary objects that can support hazard recognition and downstream safety-related decision-making}. \textcolor{black}{Moving targets include pedestrians, vehicles, and other micromobility users \cite{Apurv2021, Sabri2024}, while stationary hazards include potholes, traffic signs, and cones \cite{Su2024}.} \textcolor{black}{
Detection accuracy alone does not establish collision reduction because safety also depends on sensing latency, false alarms, missed hazards, warning logic, and rider response.}

\subsection{Object Tracking \& Trajectory Estimation}


\textcolor{black}{Object tracking and trajectory estimation use current and past positions to anticipate future motion and potential collisions. In micromobility, these methods can also be applied to advanced collision prevention systems. For instance, \cite{Yamaguchi2022} presents a bicycle-based system that predicts pedestrian trajectories to anticipate potential collisions. }\textcolor{black}{
Unlike systems that rely only on current positions, this approach forecasts whether and when a pedestrian may cross the bicycle's path, allowing warnings to be generated earlier. Real-world safety benefits additionally depend on warning latency, false-alarm rates, rider response, and deployment conditions.}


\subsection{Other Applications}

In addition to the applications mentioned, \ac{ml} methods have also been employed in micromobility for other purposes.

\begin{itemize}

    \item[\textit{a.}] \textit{Battery Health Prediction:} 
    \textcolor{black}{
    Battery health prediction can support maintenance and charging decisions in electric micromobility by identifying degradation patterns and potential battery issues \cite{Flores2023}, but its practical value depends on whether improved predictions lead to measurable benefits such as lower maintenance costs, improved battery utilisation, or safer operation.}

    \item[\textit{b.}] \textit{Trip Mode \& Purpose Analysis:} 
    \textcolor{black}{
    \ac{ml}-based analysis of travel behaviour, trip patterns, and trip purposes can provide useful information for vehicle deployment and service planning \cite{Zhang2023, Xing2020}. These applications are particularly relevant for understanding how micromobility is used within broader travel behaviour, although behavioural inference alone does not demonstrate improvements in service efficiency or user satisfaction.}

    \item[\textit{c.}] \textit{User Safety Concerns:} 
    \ac{ml} methods have been used to identify factors associated with e-bike crash injury severity \cite{Wang2021_1}, assess cyclist road-safety conditions \cite{DeBock2022}, and examine built-environment effects on cycling safety \cite{Costa2024}. \textcolor{black}{
    These analyses can support risk identification and prioritisation, but their value for safety improvement ultimately depends on whether subsequent interventions reduce actual crash risk.}

    \item[\textit{d.}] \textit{Infrastructure Improvements:} 
    \textcolor{black}{
    Data-driven and \ac{ml}-based approaches can help identify high-risk locations and support the assessment of routes or infrastructure \cite{Tamagusko2023}. Such information can guide planning decisions, although the effectiveness of the resulting infrastructure or routing interventions requires evaluation under real operating conditions.}

    \item[\textit{e.}] \textit{Deployment Considerations:} 
    Practical deployment of \ac{ml} models in micromobility often requires operation on edge or embedded platforms. Lightweight neural architectures \cite{Sandler2018, zhang2018shufflenet, iandola2016squeezenet} and model-compression techniques such as pruning and quantisation \cite{Roth2024} have therefore been explored for resource-constrained applications \cite{Kaundanya2024}. \textcolor{black}{
    Deployment feasibility should additionally be assessed in terms of target-device latency, memory and energy requirements, and task-specific failure behaviour.}

\end{itemize}

\begin{table*}[ht]
    \caption{Applications.}
    \vspace{-0.1in}
    \label{table: applications}
    \scriptsize
    \begin{adjustbox}{width=\textwidth}
    \begin{tabularx}{\linewidth}{@{\extracolsep{\fill}}X X c}
        \toprule
        \textbf{Application} & \textbf{Methods used} & \textbf{References} \\
        \midrule
        Personal mobility inattentive driving monitoring system & Short-time Fourier Transform, Wavelet Transform, CNN & \cite{Kim2022} \\
        \midrule
        Accident prevention on E-bikes for micromobility rider & CNN (object detection), Optical Flow, Object Tracking using CV & \cite{Subramanyam2023} \\
        \midrule
        Bicycle-based collision prevention system using pedestrian trajectory prediction & CNN (object detection), Object Tracking using CV, TTC using depth camera & \cite{Yamaguchi2022} \\
        \midrule
        Object-localisation and classification for resource-constrained platforms such as micromobility & CNN (Novel Cell-wise segmentation) & \cite{Su2024} \\
        \midrule
        Shared mobility demand prediction & ARIMA, STGCN, Conv-LSTM, GRU, RNN, LSTM, GCN, LR, RF, MLP, etc & \cite{Guidon2019, Zhou2022, Saum2020, Phithakkitnukooon2021, Boonjubut2022, Lee2022, Luo2021, Xiao2020, Saum2024, PelezRodrguez2024, Ma2020, Hua2020, Cuong2024, Yang2020, E2020, Shi2023} \\
        \midrule
        Shared mobility destination prediction & Conv-LSTM, GCN, RNN, LSTM, GRU, CNN, XGB, ARIMA, etc & \cite{Liang2022, Nawaz2020, Zong2019, Wang2021, Miao2021, Jiang2019, Liu2019, Li2021, Chang2020, Tsiligkaridis2022, Zhao2022, Zhang2023_1} \\
        \midrule
        E-mobility energy consumption prediction & RF, GB, ANN, kNN, MLR, LGBM, XGB, etc & \cite{Ouf2023, Ina2022, Gioldasis2024} \\
        \midrule
        Shared mobility route planning with constraints & LR, K-means, SVM, XGB, MLP, LSTM, etc & \cite{Scott2021, Masoud2023, Li2024} \\
        \bottomrule
    \end{tabularx}
    \end{adjustbox}
    \vspace{-0.2in}
\end{table*}

\color{black}


Predictive performance should be distinguished from downstream operational and safety outcomes, which depend on how model outputs are integrated into real-world decision-making. 
These outcomes are also influenced by data quality, computational resources, and the privacy and security requirements associated with mobility trajectories, user information, and sensor data.

\color{black}


\section{Research Directions \& Challenges}
\label{sec: direction}

\color{black}


Our study reviewed recent micromobility datasets, \ac{ml} techniques, and their applications. \textcolor{black}{
The reviewed literature reveals limitations in data coverage, model evaluation, and application-level evidence that constrain the generalisability and practical use of current approaches.} These challenges are discussed from the perspectives of datasets, techniques, and applications.

\color{black}

\subsection{Datasets}

\color{black}

A key limitation is the temporal and geographical concentration of available micromobility datasets. Among the 21 datasets presented in \autoref{fig: tabular dataset}, 12 span two years or less, while 18 are associated with a single city or location, 2 cover multiple geographical areas, and 1 has unspecified geographical coverage. Among the 95 included studies presented in the \textit{Supplementary Material}, 69 used a single-city or single-area setting, 20 covered multiple locations, and 6 had unreported or inapplicable locations. These counts describe geographical coverage. Only two studies explicitly evaluated cross-city transfer, including one that used target-city pilot data. These characteristics limit long-term and geographical generalisation. In the short term, dataset publishers and operators should improve reporting and data comparability, while longer-term progress requires comparable longitudinal datasets across cities and operators.


Data also remain fragmented across vehicle types, sensing modalities, and contextual layers. Existing work has begun to combine heterogeneous spatial relationships and contextual information for micromobility forecasting \cite{Xu2023_2}, showing that this direction is under-explored rather than absent. Future datasets should better align GPS trajectories, inertial or vibration signals, images, infrastructure attributes, weather, and public transport information within reusable data structures.


A further micromobility-specific challenge is the limited transferability of datasets developed for other mobility domains because of different sensing perspectives and motion dynamics. Autonomous-driving image datasets, for example, differ from micromobility data in camera placement, vehicle motion, and operating environment. \textcolor{black}{
Dedicated micromobility datasets should therefore document sensing position, camera configuration, road environment, and recording conditions, while longer-term efforts should expand coverage across devices, cities, and infrastructure settings.}


\color{black}

\subsection{Techniques}

\color{black}

A central technical challenge is modelling micromobility spatiotemporal dynamics \textcolor{black}{
across different spatial structures and temporal scales. Model performance can depend strongly on the prediction target, dataset, and graph construction, while multi-graph approaches have begun to represent heterogeneous spatial relationships explicitly \cite{Chai2018, Xu2023_2}. 
Comparability can be improved by evaluating simpler temporal and tree-based baselines under consistent inputs, data splits, and forecasting horizons, and by reporting graph and attention ablations.}



\textcolor{black}{
Generalisation is challenged by differences across cities, operators, seasons, and devices. Evaluation in multiple cities does not demonstrate transfer to an unseen city or operator when each location is trained and tested separately \cite{Xu2023_2}. Generalisation also varies across riders. In \cite{Kim2022}, mean within-rider accuracy was 91.82\%, compared with 43.07\% when a rider was held out. Strong personalised performance therefore did not carry over to an unseen rider in that study. 
Changes in weather, infrastructure, policy, fleet operation, and user behaviour may require model monitoring and updating \cite{Lyu2024}. 
Short-term robustness tests should use chronological splits, held-out locations, and missing inputs, followed by broader cross-city and cross-operator evaluation.}

\color{black}


Multimodal and multi-representation modelling is also under-explored rather than absent. Existing studies combine complementary signal representations for rider-attentiveness monitoring \cite{Kim2022} and multiple spatial graphs for scooter-demand forecasting \cite{Xu2023_2}. 
Future evaluation should address missing inputs, asynchronous sampling, and information leakage through modality and graph ablations and by fitting data-driven preprocessing using training data only. Broader integration with infrastructure and public transport data will require longer-term data-sharing arrangements.

\color{black}

Finally, resource-constrained learning is \textcolor{black}{
particularly relevant to micromobility safety applications requiring on-device or near-real-time inference. Lightweight lane-recognition models have already been investigated \cite{Kaundanya2024}, but deployment feasibility also depends on target hardware, numerical precision, memory use, energy consumption, and end-to-end latency. 
Longer-term field studies are needed to assess reliability under changing road, weather, and operating conditions.}


\color{black}

\subsection{Applications}

\color{black}


A central application-level challenge is the gap between predictive performance and system-level outcomes. Many studies focus on forecasting or classification accuracy, while equitable access, public transport integration, operating costs, and service reliability require decision-level evaluation.


Fairness-aware demand prediction has already incorporated spatial disparities into learning objectives \cite{Yan2020}, while more recent work published after the review period has combined cross-modal graph learning with equity-aware prediction \cite{Yang2025}. 
In micromobility, \ac{ml} can support fairness by identifying spatial disparities in demand and service availability, incorporating fairness constraints into allocation or rebalancing, and examining whether errors are systematically larger in underserved areas. However, similar prediction errors do not necessarily imply equitable access because observed demand may itself reflect unequal supply, affordability, or connectivity.


Public transport integration is another under-explored direction. \ac{ml} can combine micromobility demand with public transport usage and contextual information to support first- and last-mile forecasting and vehicle allocation. Recent work published after the review period has integrated subway usage with shared micromobility demand forecasting \cite{Belkessa2026}. 
Evaluation should extend beyond forecasting accuracy to outcomes such as vehicle availability near transport stops, unmet demand, transfer opportunities, and accessibility across population groups.


Privacy is particularly important because micromobility datasets can contain fine-grained trajectories, timestamps, and behavioural information. Recent work published after the review period has investigated federated demand prediction without directly pooling raw data \cite{Zhu2025}. 
Federated learning does not by itself guarantee privacy, so evaluation should specify the threat model, shared information, privacy mechanisms, and trade-offs among privacy, predictive utility, and communication cost.


Operational and economic decision-making also remains under-explored. Relevant studies have examined incentive-based rebalancing, dynamic pricing, and prediction-informed rebalancing \cite{Singla2015, Shi2024, Saum2024}. 
These approaches provide different levels of evidence, ranging from observed user responses to optimisation and simulation under modelled assumptions. 
Short-term evaluation should report operational indicators such as vehicle availability, unmet requests, rebalancing distance, utilisation, revenue, and cost under defined constraints. 
Longer-term studies should assess whether pricing, incentives, rebalancing, and infrastructure interventions produce sustained changes in access, mode choice, emissions, and safety.

Overall, future micromobility applications should be evaluated not only by model accuracy but also by whether they improve the intended service, operational, or societal outcome.

\color{black}

\section{Conclusion}
\label{sec: conclusion}

\textcolor{black}{
This systematic review examined 95 studies on micromobility datasets, \ac{ml} techniques, and applications, including demand prediction, trip destination forecasting, energy management, route planning, and rider safety. 
The reviewed literature shows that the suitability and reported performance of \ac{ml} methods depend strongly on the available data, modelling assumptions, and evaluation settings. Differences in spatial and temporal coverage, input features, data splits, aggregation levels, and prediction horizons therefore limit direct comparison across studies. 
Evidence of geographical generalisation also remains limited. Although several studies were conducted across multiple cities or geographical areas, only two explicitly evaluated cross-city transfer or generalisation, including one that used target-city pilot data. Evaluation in multiple locations should therefore not be interpreted as evidence of transfer to an unseen city or operator when models are trained and tested separately within each location. 
Improvements in predictive performance also do not necessarily translate into better operational, safety, or societal outcomes. Lower forecasting or classification error alone does not demonstrate improved vehicle availability, reduced rebalancing costs, more equitable access, or improved rider safety. Future work should therefore adopt more comparable evaluation protocols, test robustness and transfer across cities and operators, and assess whether model-informed decisions deliver measurable practical benefits under realistic operating conditions.}

This review still has several limitations. It mainly focuses on open-access datasets, while private datasets may contain additional information that is not represented in the reviewed resources. Moreover, while the review provides technical comparisons of datasets, models, and applications, it \textcolor{black}{
does not systematically examine} the effects of policy, regulation, user acceptance, or broader institutional conditions on micromobility deployment. \textcolor{black}{
Operations-oriented decision-making problems, including pricing, incentives, and operator key performance indicators, are also less consistently represented in the reviewed technical literature.} \textcolor{black}{
The search applied publication-date restrictions without additional search-field restrictions, but the exact database-specific syntax used during the original searches was not retained, which limits complete replication of the search procedure.} Finally, although large language models and generative artificial intelligence have recently attracted substantial attention, their use in micromobility-focused \ac{ml} studies remains relatively sparse and heterogeneous. They are therefore not treated as a core technique category in this review and remain a direction for future work.

\section*{Acknowledgement} 
\label{sec: acknowledgement}

This publication has emanated from research conducted with the financial support of Taighde Éireann — Research Ireland under Grant number \textit{21/FFP-P/10266} and \textit{SFI/12/RC/2289\_P2}.


\bibliographystyle{IEEEtran}
\bibliography{reference}

@article{Apurv2021,
  title = {Detection of E-scooter Riders in Naturalistic Scenes},
  DOI = {10.48550/ARXIV.2111.14060},
  journal={arXiv preprint arXiv:2111.14060},
  publisher = {arXiv},
  author = {Apurv,  Kumar and Tian,  Renran and Sherony,  Rini},
  year={2021}
}

@article{Erdeli2023,
  title = {Dataset for multimodal transport analytics of smartphone users - Collecty},
  volume = {50},
  ISSN = {2352-3409},
  DOI = {10.1016/j.dib.2023.109481},
  journal = {Data in Brief},
  publisher = {Elsevier BV},
  author = {Erdelić,  Martina and Erdelić,  Tomislav and Carić,  Tonči},
  year = {2023},
  month = oct,
  pages = {109481}
}

@article{Su2024,
  title = {YOLIC: An efficient method for object localization and classification on edge devices},
  volume = {147},
  ISSN = {0262-8856},
  DOI = {10.1016/j.imavis.2024.105095},
  journal = {Image and Vision Computing},
  publisher = {Elsevier BV},
  author = {Su,  Kai and Tomioka,  Yoichi and Zhao,  Qiangfu and Liu,  Yong},
  year = {2024},
  month = jul,
  pages = {105095}
}

@inproceedings{Nienaber2015_1,
  doi = {10.13140/RG.2.1.3121.8408},
  author = {{S Nienaber} and {Mj Booysen} and {Rs Kroon}},
  language = {en},
  title = {Detecting Potholes Using Simple Image Processing Techniques and Real-World Footage},
  publisher = {SATC},
  booktitle = {34th Annual Southern African Transport Conference 2015},
  year = {2015}
}

@inproceedings{Nienaber2015_2,
  title = {A Comparison of Low-Cost Monocular Vision Techniques for Pothole Distance Estimation},
  DOI = {10.1109/ssci.2015.69},
  booktitle = {2015 IEEE Symposium Series on Computational Intelligence},
  publisher = {IEEE},
  author = {Nienaber,  S. and Kroon,  R.S. and Booysen,  M.J.},
  year = {2015},
  month = dec 
}

@misc{Lyft2019,
  title = {Bay Wheels System Data},
  author = {{Lyft Inc.}},
  year = {2017},
  howpublished = {\url{https://www.lyft.com/bikes/bay-wheels/system-data}},
  note = {Accessed: 2024-07-10}
}

@misc{Lyft2024,
  title = {Citi Bike System Data},
  author = {{Lyft Inc.}},
  year = {2024},
  howpublished = {\url{https://citibikenyc.com/system-data}},
  note = {Accessed: 2024-07-10}
}

@misc{CYCLANDS,
  doi = {10.5281/ZENODO.5603035},
  url = {https://zenodo.org/record/5603035},
  author = {Costa,  Miguel and Marques,  Manuel and Roque,  Carlos and Moura,  Filipe},
  title = {CYCLANDS: CYCling geo-Located AccideNts,  their Details and Severities},
  publisher = {Zenodo},
  year = {2021},
  copyright = {Open Access}
}

@misc{Norfolk2019,
  title = {Micromobility: Electric Scooters and Bikes},
  author = {{City of Norfolk}},
  year = {2024},
  howpublished = {\url{https://data.norfolk.gov/Government/Micromobility-Electric-Scooters-and-Bikes-/wqxq-hhe6/about_data}},
  note = {Accessed: 2024-07-10}
}

@misc{Madrid2022,
  title = {BiciMAD E-bike Daily Usage Record},
  author = {{City of Madrid}},
  year = {2022},
  howpublished = {\url{https://datos.madrid.es/sites/v/index.jsp?vgnextoid=6d8bdae2be63c410VgnVCM1000000b205a0aRCRD&vgnextchannel=374512b9ace9f310VgnVCM100000171f5a0aRCRD}},
  note = {Accessed: 2024-07-24}
}

@misc{Barcelona2018,
  title = {Bicing Shared Bike Usage Record},
  author = {{City of Barcelona}},
  year = {2018},
  howpublished = {\url{https://opendata-ajuntament.barcelona.cat/data/en/dataset/us-del-servei-bicing}},
  note = {Accessed: 2024-07-24}
}

@misc{Seoul2018,
  title = {Seoul Bike Sharing Rental History},
  author = {{City of Seoul}},
  year = {2017},
  howpublished = {\url{https://data.seoul.go.kr/dataList/OA-15182/F/1/datasetView.do}},
  note = {Accessed: 2024-07-20}
}

@misc{Austin2018,
  title = {Austin Shared Mobility Vehicle Trips},
  author = {{City of Austin}},
  year = {2019},
  howpublished = {\url{https://data.austintexas.gov/Transportation-and-Mobility/Shared-Micromobility-Vehicle-Trips-2018-2022-/7d8e-dm7r/}},
  note = {Accessed: 2024-07-29}
}

@misc{Biketown2018,
  title = {Biketown Trip Data},
  author = {{Biketown}},
  year = {2018},
  howpublished = {\url{https://biketownpdx.com/system-data}},
  note = {Accessed: 2024-07-20}
}

@misc{Toronto2017,
  title = {Bike Share Toronto Ridership Data},
  author = {{City of Toronto}},
  year = {2018},
  howpublished = {\url{https://open.toronto.ca/dataset/bike-share-toronto-ridership-data}},
  note = {Accessed: 2024-07-20}
}

@misc{Chicago2022,
  title = {E-Scooter Trip Data},
  author = {{City of Chicago}},
  year = {2022},
  howpublished = {\url{https://data.cityofchicago.org/Transportation/E-Scooter-Trips/2i5w-ykuw}},
  note = {Accessed: 2024-07-20}
}

@misc{Moby2020,
  title = {Moby Bikes Bikeshare},
  author = {{City of Dublin}},
  year = {2020},
  howpublished = {\url{https://data.gov.ie/dataset/moby-bikes}},
  note = {Accessed: 2024-07-20}
}

@misc{Minneapolis2019,
  title = {Motorized Foot Scooter Trips},
  author = {{City of Minneapolis}},
  year = {2019},
  howpublished = {\url{https://opendata.minneapolismn.gov/datasets/cityoflakes::motorized-foot-scooter-trips-may-2019/about}},
  note = {Accessed: 2024-07-23}
}

@article{Guidon2019,
  title = {Electric Bicycle-Sharing: A New Competitor in the Urban Transportation Market? An Empirical Analysis of Transaction Data},
  volume = {2673},
  ISSN = {2169-4052},
  DOI = {10.1177/0361198119836762},
  number = {4},
  journal = {Transportation Research Record: Journal of the Transportation Research Board},
  publisher = {SAGE Publications},
  author = {Guidon,  Sergio and Becker,  Henrik and Dediu,  Horace and Axhausen,  Kay W.},
  year = {2019},
  month = mar,
  pages = {15–26}
}

@article{Zhou2022,
  title = {Spatiotemporal characteristics analysis of commuting by shared electric bike: A case study of Ningbo,  China},
  volume = {362},
  ISSN = {0959-6526},
  DOI = {10.1016/j.jclepro.2022.132337},
  journal = {Journal of Cleaner Production},
  publisher = {Elsevier BV},
  author = {Zhou,  Xizhen and Ji,  Yanjie and Yuan,  Yidan and Zhang,  Fan and An,  Qinhe},
  year = {2022},
  month = aug,
  pages = {132337}
}

@inproceedings{Saum2020,
  title = {Short-Term Demand and Volatility Prediction of Shared Micro-Mobility: a case study of e-scooter in Thammasat University},
  DOI = {10.1109/fists46898.2020.9264852},
  booktitle = {2020 Forum on Integrated and Sustainable Transportation Systems (FISTS)},
  publisher = {IEEE},
  author = {Saum,  Narith and Sugiura,  Satoshi and Piantanakulchai,  Mongkut},
  year = {2020},
  month = nov 
}

@article{Phithakkitnukooon2021,
  title = {Predicting Spatiotemporal Demand of Dockless E-Scooter Sharing Services with a Masked Fully Convolutional Network},
  volume = {10},
  ISSN = {2220-9964},
  DOI = {10.3390/ijgi10110773},
  number = {11},
  journal = {ISPRS International Journal of Geo-Information},
  publisher = {MDPI AG},
  author = {Phithakkitnukooon,  Santi and Patanukhom,  Karn and Demissie,  Merkebe Getachew},
  year = {2021},
  month = nov,
  pages = {773}
}

@article{Boonjubut2022,
  title = {Accuracy of Hourly Demand Forecasting of Micro Mobility for Effective Rebalancing Strategies},
  volume = {30},
  ISSN = {2450-5781},
  DOI = {10.2478/mspe-2022-0031},
  number = {3},
  journal = {Management Systems in Production Engineering},
  publisher = {Walter de Gruyter GmbH},
  author = {Boonjubut,  Kanokporn and Hasegawa,  Hiroshi},
  year = {2022},
  month = jul,
  pages = {246–252}
}

@article{Lee2022,
  title = {Evaluation of Spatial and Temporal Performance of Deep Learning Models for Travel Demand Forecasting: Application to Bike-Sharing Demand Forecasting},
  volume = {2022},
  ISSN = {0197-6729},
  DOI = {10.1155/2022/5934670},
  journal = {Journal of Advanced Transportation},
  publisher = {Hindawi Limited},
  author = {Lee,  Jaehyung and Kim,  Jinhee},
  editor = {González-Ramírez,  Rosa G.},
  year = {2022},
  month = jun,
  pages = {1–13}
}

@article{Luo2021,
  title = {Predicting Travel Demand of a Docked Bikesharing System Based on LSGC-LSTM Networks},
  volume = {9},
  ISSN = {2169-3536},
  DOI = {10.1109/access.2021.3062778},
  journal = {IEEE Access},
  publisher = {Institute of Electrical and Electronics Engineers (IEEE)},
  author = {Luo,  Jing and Zhou,  Dai and Han,  Zhaolong and Xiao,  Guangnian and Tan,  Yunlong},
  year = {2021},
  pages = {92189–92203}
}

@article{Xiao2020,
  title = {Demand prediction for a public bike sharing program based on spatio-temporal graph convolutional networks},
  volume = {80},
  ISSN = {1573-7721},
  DOI = {10.1007/s11042-020-08803-y},
  number = {15},
  journal = {Multimedia Tools and Applications},
  publisher = {Springer Science and Business Media LLC},
  author = {Xiao,  Guangnian and Wang,  Ruinan and Zhang,  Chunqin and Ni,  Anning},
  year = {2020},
  month = mar,
  pages = {22907–22925}
}

@article{Liang2022,
  title = {Combining Individual Travel Preferences Into Destination Prediction: A Multi-Module Deep Learning Network},
  volume = {23},
  ISSN = {1558-0016},
  DOI = {10.1109/tits.2021.3128153},
  number = {8},
  journal = {IEEE Transactions on Intelligent Transportation Systems},
  publisher = {Institute of Electrical and Electronics Engineers (IEEE)},
  author = {Liang,  Jian and Tang,  Jinjun and Liu,  Fang and Wang,  Yinhai},
  year = {2022},
  month = aug,
  pages = {13782–13793}
}

@inbook{Nawaz2020,
  title = {SSMDL: Semi-supervised Multi-task Deep Learning for Transportation Mode Classification and Path Prediction with GPS Trajectories},
  ISBN = {9783030602901},
  ISSN = {1611-3349},
  DOI = {10.1007/978-3-030-60290-1_31},
  booktitle = {Lecture Notes in Computer Science},
  publisher = {Springer International Publishing},
  author = {Nawaz,  Asif and Huang,  Zhiqiu and Wang,  Senzhang},
  year = {2020},
  pages = {391–405}
}

@article{Zong2019,
  title = {Trip destination prediction based on multi-day GPS data},
  volume = {515},
  ISSN = {0378-4371},
  DOI = {10.1016/j.physa.2018.09.090},
  journal = {Physica A: Statistical Mechanics and its Applications},
  publisher = {Elsevier BV},
  author = {Zong,  Fang and Tian,  Yongda and He,  Yanan and Tang,  Jinjun and Lv,  Jianyu},
  year = {2019},
  month = feb,
  pages = {258–269}
}

@article{Wang2021,
  title = {An Attention-Based Deep Learning Framework for Trip Destination Prediction of Sharing Bike},
  volume = {22},
  ISSN = {1558-0016},
  DOI = {10.1109/tits.2020.3008935},
  number = {7},
  journal = {IEEE Transactions on Intelligent Transportation Systems},
  publisher = {Institute of Electrical and Electronics Engineers (IEEE)},
  author = {Wang,  Wei and Zhao,  Xiaofeng and Gong,  Zhiguo and Chen,  Zhikui and Zhang,  Ning and Wei,  Wei},
  year = {2021},
  month = jul,
  pages = {4601–4610}
}

@article{Miao2021,
  title = {Deep learning based origin-destination prediction via contextual information fusion},
  volume = {81},
  ISSN = {1573-7721},
  DOI = {10.1007/s11042-020-10492-6},
  number = {9},
  journal = {Multimedia Tools and Applications},
  publisher = {Springer Science and Business Media LLC},
  author = {Miao,  Hao and Fei,  Yan and Wang,  Senzhang and Wang,  Fang and Wen,  Danyan},
  year = {2021},
  month = jan,
  pages = {12029–12045}
}

@article{Jiang2019,
  title = {A Destination Prediction Network Based on Spatiotemporal Data for Bike-Sharing},
  volume = {2019},
  ISSN = {1099-0526},
  DOI = {10.1155/2019/7643905},
  journal = {Complexity},
  publisher = {Hindawi Limited},
  author = {Jiang,  Jian and Lin,  Fei and Fan,  Jin and Lv,  Hang and Wu,  Jia},
  year = {2019},
  month = jan,
  pages = {1–14}
}

@article{Liu2019,
  title = {A Two-Stage Destination Prediction Framework of Shared Bicycles Based on Geographical Position Recommendation},
  volume = {11},
  ISSN = {1941-1197},
  DOI = {10.1109/mits.2018.2884517},
  number = {1},
  journal = {IEEE Intelligent Transportation Systems Magazine},
  publisher = {Institute of Electrical and Electronics Engineers (IEEE)},
  author = {Liu,  Yang and Jia,  Ruo and Xie,  Xue and Liu,  Zhiyuan},
  year = {2019},
  pages = {42–47}
}

@inproceedings{Flores2023,
  title = {Leveraging IoT and TinyML for Smart Battery Management in Electric Bicycles},
  DOI = {10.1109/siot60039.2023.10390169},
  booktitle = {2023 Symposium on Internet of Things (SIoT)},
  publisher = {IEEE},
  author = {Flores,  Thommas and Andrade,  Matheus and Medeiros,  Morsinaldo and Amaral,  Miguel and Silva,  Marianne and Silva,  Ivanovitch},
  year = {2023},
  month = oct 
}

@inproceedings{Ouf2023,
  title = {E-Bike Energy Needs Estimation based on Route Characteristics and Rider Behavior},
  DOI = {10.1109/icicis58388.2023.10391196},
  booktitle = {2023 Eleventh International Conference on Intelligent Computing and Information Systems (ICICIS)},
  publisher = {IEEE},
  author = {Ouf,  Karim and Soubra,  Hassan and Mazhr,  Ahmed},
  year = {2023},
  month = nov 
}

@article{Ina2022,
  title = {Estimation of Postal Service Delivery Time and Energy Cost with E-Scooter by Machine Learning Algorithms},
  volume = {12},
  ISSN = {2076-3417},
  DOI = {10.3390/app122312266},
  number = {23},
  journal = {Applied Sciences},
  publisher = {MDPI AG},
  author = {İna\c{c},  Hakan and Ay\"{o}zen,  Yunus Emre and Atalan,  Abdulkadir and D\"{o}nmez,  Cem undefinedağrı},
  year = {2022},
  month = nov,
  pages = {12266}
}

@inproceedings{Yamaguchi2022,
  title = {Bicycle-based collision prevention system using pedestrian trajectory prediction},
  DOI = {10.1109/candarw57323.2022.00013},
  booktitle = {2022 Tenth International Symposium on Computing and Networking Workshops (CANDARW)},
  publisher = {IEEE},
  author = {Yamaguchi,  Nobuhiko and Tokumaru,  Hiroki and Fukuda,  Osamu and Okumura,  Hiroshi and Yeoh,  Wen Liang},
  year = {2022},
  month = nov 
}

@inproceedings{Subramanyam2023,
  title = {Accident Prediction on E-Bikes Using Computer Vision},
  DOI = {10.1109/bigdataservice58306.2023.00040},
  booktitle = {2023 IEEE Ninth International Conference on Big Data Computing Service and Applications (BigDataService)},
  publisher = {IEEE},
  author = {Subramanyam,  Rohith Puvvala and Naik,  Abhay and Suresh,  Mahima Agumbe},
  year = {2023},
  month = jul 
}

@article{Kim2022,
  title = {Safety monitoring system of personal mobility driving using deep learning},
  volume = {9},
  ISSN = {2288-5048},
  DOI = {10.1093/jcde/qwac061},
  number = {4},
  journal = {Journal of Computational Design and Engineering},
  publisher = {Oxford University Press (OUP)},
  author = {Kim,  Eunji and Ryu,  Hanyoung and Oh,  Hyunji and Kang,  Namwoo},
  year = {2022},
  month = jul,
  pages = {1397–1409}
}

@article{Ahmed2022,
  title = {Survey of Machine Learning Methods Applied to Urban Mobility},
  volume = {10},
  ISSN = {2169-3536},
  DOI = {10.1109/access.2022.3159668},
  journal = {IEEE Access},
  publisher = {Institute of Electrical and Electronics Engineers (IEEE)},
  author = {Ahmed,  Dina Bousdar and Diaz,  Estefania Munoz},
  year = {2022},
  pages = {30349–30366}
}

@article{Masoud2023,
  title = {A Hybrid K-Means and Particle Swarm Optimization Technique for Solving the Rechargeable E-Scooters Problem},
  volume = {11},
  ISSN = {2169-3536},
  DOI = {10.1109/access.2023.3336810},
  journal = {IEEE Access},
  publisher = {Institute of Electrical and Electronics Engineers (IEEE)},
  author = {Masoud,  Mahmoud},
  year = {2023},
  pages = {132472–132482}
}

@article{Saum2024,
  title = {Optimizing Shared E-Scooter Operations Under Demand Uncertainty: A Framework Integrating Machine Learning and Optimization Techniques},
  volume = {12},
  ISSN = {2169-3536},
  DOI = {10.1109/access.2024.3365947},
  journal = {IEEE Access},
  publisher = {Institute of Electrical and Electronics Engineers (IEEE)},
  author = {Saum,  Narith and Sugiura,  Satoshi and Piantanakulchai,  Mongkut},
  year = {2024},
  pages = {26957–26977}
}

@inproceedings{Li2024,
  series = {WSDM ’24},
  title = {Real-time E-bike Route Planning with Battery Range Prediction},
  DOI = {10.1145/3616855.3635696},
  booktitle = {Proceedings of the 17th ACM International Conference on Web Search and Data Mining},
  publisher = {ACM},
  author = {Li,  Zhao and Ren,  Guoqi and Gu,  Yongchun and Zhou,  Siwei and Liu,  Xuanwu and Huang,  Jiaming and Li,  Ming},
  year = {2024},
  month = mar,
  collection = {WSDM ’24}
}

@inproceedings{Kaundanya2024,
  title = {Using Attention Mechanisms in Compact CNN Models for Improved Micromobility Safety Through Lane Recognition},
  DOI = {10.5220/0012600300003702},
  booktitle = {Proceedings of the 10th International Conference on Vehicle Technology and Intelligent Transport Systems},
  publisher = {SCITEPRESS - Science and Technology Publications},
  author = {Kaundanya,  Chinmaya and Cesar,  Paulo and Cronin,  Barry and Fleury,  Andrew and Liu,  Mingming and Little,  Suzanne},
  year = {2024}
}

@article{Gioldasis2024,
  title = {Usage factors influencing e-scooter energy consumption: An empirical investigation},
  volume = {452},
  ISSN = {0959-6526},
  DOI = {10.1016/j.jclepro.2024.142165},
  journal = {Journal of Cleaner Production},
  publisher = {Elsevier BV},
  author = {Gioldasis,  Christos and Christoforou,  Zoi and Katsiadrami,  Aikaterini},
  year = {2024},
  month = may,
  pages = {142165}
}

@article{PelezRodrguez2024,
  title = {Bike sharing and cable car demand forecasting using machine learning and deep learning multivariate time series approaches},
  volume = {238},
  ISSN = {0957-4174},
  DOI = {10.1016/j.eswa.2023.122264},
  journal = {Expert Systems with Applications},
  publisher = {Elsevier BV},
  author = {Peláez-Rodríguez,  César and Pérez-Aracil,  Jorge and Fister,  Dušan and Torres-López,  Ricardo and Salcedo-Sanz,  Sancho},
  year = {2024},
  month = mar,
  pages = {122264}
}

@article{Ma2020,
  title = {A comparison in travel patterns and determinants of user demand between docked and dockless bike-sharing systems using multi-sourced data},
  volume = {139},
  ISSN = {0965-8564},
  DOI = {10.1016/j.tra.2020.06.022},
  journal = {Transportation Research Part A: Policy and Practice},
  publisher = {Elsevier BV},
  author = {Ma,  Xinwei and Ji,  Yanjie and Yuan,  Yufei and Van Oort,  Niels and Jin,  Yuchuan and Hoogendoorn,  Serge},
  year = {2020},
  month = sep,
  pages = {148–173}
}

@article{Hua2020,
  title = {Estimating the parking demand of free-floating bike sharing: A journey-data-based study of Nanjing,  China},
  volume = {244},
  ISSN = {0959-6526},
  DOI = {10.1016/j.jclepro.2019.118764},
  journal = {Journal of Cleaner Production},
  publisher = {Elsevier BV},
  author = {Hua,  Mingzhuang and Chen,  Xuewu and Zheng,  Shujie and Cheng,  Long and Chen,  Jingxu},
  year = {2020},
  month = jan,
  pages = {118764}
}

@article{Cuong2024,
  title = {Analyzing Shared Bike Usage Through Graph-Based Spatio-Temporal Modeling},
  volume = {5},
  ISSN = {2687-7813},
  DOI = {10.1109/ojits.2024.3350213},
  journal = {IEEE Open Journal of Intelligent Transportation Systems},
  publisher = {Institute of Electrical and Electronics Engineers (IEEE)},
  author = {Cuong,  Dinh Viet and Ngo,  Vuong M. and Cappellari,  Paolo and Roantree,  Mark},
  year = {2024},
  pages = {115–131}
}

@article{Abduljabbar2021,
  title = {The role of micro-mobility in shaping sustainable cities: A systematic literature review},
  volume = {92},
  ISSN = {1361-9209},
  DOI = {10.1016/j.trd.2021.102734},
  journal = {Transportation Research Part D: Transport and Environment},
  publisher = {Elsevier BV},
  author = {Abduljabbar,  Rusul L. and Liyanage,  Sohani and Dia,  Hussein},
  year = {2021},
  month = mar,
  pages = {102734}
}

@article{Teusch2023,
  title = {A Systematic Literature Review on Machine Learning in Shared Mobility},
  volume = {4},
  ISSN = {2687-7813},
  DOI = {10.1109/ojits.2023.3334393},
  journal = {IEEE Open Journal of Intelligent Transportation Systems},
  publisher = {Institute of Electrical and Electronics Engineers (IEEE)},
  author = {Teusch,  Julian and Gremmel,  Jan Niklas and Koetsier,  Christian and Johora,  Fatema Tuj and Sester,  Monika and Woisetschl\"{a}ger,  David M. and M\"{u}ller,  J\"{o}rg P.},
  year = {2023},
  pages = {870–899}
}

@article{Mitropoulos2023,
  title = {E-scooter micromobility systems: Review of attributes and impacts},
  volume = {21},
  ISSN = {2590-1982},
  DOI = {10.1016/j.trip.2023.100888},
  journal = {Transportation Research Interdisciplinary Perspectives},
  publisher = {Elsevier BV},
  author = {Mitropoulos,  Lambros and Stavropoulou,  Eirini and Tzouras,  Panagiotis and Karolemeas,  Christos and Kepaptsoglou,  Konstantinos},
  year = {2023},
  month = sep,
  pages = {100888}
}

@inbook{Milakis2020,
  title = {Is micro-mobility sustainable? An overview of implications for accessibility,  air pollution,  safety,  physical activity and subjective wellbeing},
  ISBN = {9781789900460},
  DOI = {10.4337/9781789900477.00030},
  booktitle = {Handbook of Sustainable Transport},
  publisher = {Edward Elgar Publishing},
  author = {Milakis,  Dimitris and Gedhardt,  Laura and Ehebrecht,  Daniel and Lenz,  Barbara},
  year = {2020},
  month = dec,
  pages = {180-189}
}

@inproceedings{Yan2023,
  title = {A Review on AI Algorithms for Energy Management in E-Mobility Services},
  DOI = {10.1109/cvci59596.2023.10397371},
  booktitle = {2023 7th CAA International Conference on Vehicular Control and Intelligence (CVCI)},
  publisher = {IEEE},
  author = {Yan,  Sen and Shah,  Maqsood Hussain and Li,  Ji and O’Connor,  Noel and Liu,  Mingming},
  year = {2023},
  month = oct 
}

@article{Yang2020,
  title = {Space-time demand cube for spatial-temporal coverage optimization model of shared bicycle system: A study using big bike GPS data},
  volume = {88},
  ISSN = {0966-6923},
  DOI = {10.1016/j.jtrangeo.2020.102861},
  journal = {Journal of Transport Geography},
  publisher = {Elsevier BV},
  author = {Yang,  Lin and Zhang,  Fayong and Kwan,  Mei-Po and Wang,  Ke and Zuo,  Zejun and Xia,  Shaotian and Zhang,  Zhiyong and Zhao,  Xinpei},
  year = {2020},
  month = oct,
  pages = {102861}
}

@article{Li2021,
  title = {Inferring the trip purposes and uncovering spatio-temporal activity patterns from dockless shared bike dataset in Shenzhen,  China},
  volume = {91},
  ISSN = {0966-6923},
  DOI = {10.1016/j.jtrangeo.2021.102974},
  journal = {Journal of Transport Geography},
  publisher = {Elsevier BV},
  author = {Li,  Shaoying and Zhuang,  Caigang and Tan,  Zhangzhi and Gao,  Feng and Lai,  Zhipeng and Wu,  Zhifeng},
  year = {2021},
  month = feb,
  pages = {102974}
}

@article{Scott2021,
  title = {Route choice of bike share users: Leveraging GPS data to derive choice sets},
  volume = {90},
  ISSN = {0966-6923},
  DOI = {10.1016/j.jtrangeo.2020.102903},
  journal = {Journal of Transport Geography},
  publisher = {Elsevier BV},
  author = {Scott,  Darren M. and Lu,  Wei and Brown,  Matthew J.},
  year = {2021},
  month = jan,
  pages = {102903}
}

@article{E2020,
  title = {Using data mining techniques for bike sharing demand prediction in metropolitan city},
  volume = {153},
  ISSN = {0140-3664},
  DOI = {10.1016/j.comcom.2020.02.007},
  journal = {Computer Communications},
  publisher = {Elsevier BV},
  author = {E,  Sathishkumar V and Park,  Jangwoo and Cho,  Yongyun},
  year = {2020},
  month = mar,
  pages = {353–366}
}

@article{Chang2020,
  title = {Understanding user’s travel behavior and city region functions from station-free shared bike usage data},
  volume = {72},
  ISSN = {1369-8478},
  DOI = {10.1016/j.trf.2020.03.018},
  journal = {Transportation Research Part F: Traffic Psychology and Behaviour},
  publisher = {Elsevier BV},
  author = {Chang,  Ximing and Wu,  Jianjun and He,  Zhengbing and Li,  Daqing and Sun,  Huijun and Wang,  Weiping},
  year = {2020},
  month = jul,
  pages = {81–95}
}

@article{Xing2020,
  title = {Exploring travel patterns and trip purposes of dockless bike-sharing by analyzing massive bike-sharing data in Shanghai,  China},
  volume = {87},
  ISSN = {0966-6923},
  DOI = {10.1016/j.jtrangeo.2020.102787},
  journal = {Journal of Transport Geography},
  publisher = {Elsevier BV},
  author = {Xing,  Yingying and Wang,  Ke and Lu,  Jian John},
  year = {2020},
  month = jul,
  pages = {102787}
}

@article{Zhang2023,
  title = {Understanding Travel Mode Choice Behavior: Influencing Factors Analysis and Prediction with Machine Learning Method},
  volume = {15},
  ISSN = {2071-1050},
  DOI = {10.3390/su151411414},
  number = {14},
  journal = {Sustainability},
  publisher = {MDPI AG},
  author = {Zhang,  Hui and Zhang,  Li and Liu,  Yanjun and Zhang,  Lele},
  year = {2023},
  month = jul,
  pages = {11414}
}

@inproceedings{Zhang2023_1,
  title = {Enhancing Urban Vehicle Destination Prediction Through Trajectory Metadata Integration and Machine Learning},
  DOI = {10.1109/acait60137.2023.10528407},
  booktitle = {2023 7th Asian Conference on Artificial Intelligence Technology (ACAIT)},
  publisher = {IEEE},
  author = {Zhang,  Ziyin and Yang,  Fei and Wang,  Li and Wang,  Zhigang and Zhou,  Tian and He,  Yansong},
  year = {2023},
  month = nov 
}

@article{Wang2021_1,
  title = {Risk factors affecting crash injury severity for different groups of e-bike riders: A classification tree-based logistic regression model},
  volume = {76},
  ISSN = {0022-4375},
  DOI = {10.1016/j.jsr.2020.12.009},
  journal = {Journal of Safety Research},
  publisher = {Elsevier BV},
  author = {Wang,  Zhengwu and Huang,  Shuai and Wang,  Jie and Sulaj,  Denisa and Hao,  Wei and Kuang,  Aiwu},
  year = {2021},
  month = feb,
  pages = {176–183}
}

@article{Liberati2009,
  title = {The PRISMA Statement for Reporting Systematic Reviews and Meta-Analyses of Studies That Evaluate Health Care Interventions: Explanation and Elaboration},
  volume = {6},
  ISSN = {1549-1676},
  DOI = {10.1371/journal.pmed.1000100},
  number = {7},
  journal = {PLoS Medicine},
  publisher = {Public Library of Science (PLoS)},
  author = {Liberati,  Alessandro and Altman,  Douglas G. and Tetzlaff,  Jennifer and Mulrow,  Cynthia and Gøtzsche,  Peter C. and Ioannidis,  John P. A. and Clarke,  Mike and Devereaux,  P. J. and Kleijnen,  Jos and Moher,  David},
  year = {2009},
  month = jul,
  pages = {e1000100}
}

@article{Rethlefsen2021,
  title = {PRISMA-S: an extension to the PRISMA Statement for Reporting Literature Searches in Systematic Reviews},
  volume = {10},
  ISSN = {2046-4053},
  DOI = {10.1186/s13643-020-01542-z},
  number = {1},
  journal = {Systematic Reviews},
  publisher = {Springer Science and Business Media LLC},
  author = {Rethlefsen,  Melissa L. and Kirtley,  Shona and Waffenschmidt,  Siw and Ayala,  Ana Patricia and Moher,  David and Page,  Matthew J. and Koffel,  Jonathan B. and Blunt,  Heather and Brigham,  Tara and Chang,  Steven and Clark,  Justin and Conway,  Aislinn and Couban,  Rachel and de Kock,  Shelley and Farrah,  Kelly and Fehrmann,  Paul and Foster,  Margaret and Fowler,  Susan A. and Glanville,  Julie and Harris,  Elizabeth and Hoffecker,  Lilian and Isojarvi,  Jaana and Kaunelis,  David and Ket,  Hans and Levay,  Paul and Lyon,  Jennifer and McGowan,  Jessie and Murad,  M. Hassan and Nicholson,  Joey and Pannabecker,  Virginia and Paynter,  Robin and Pinotti,  Rachel and Ross-White,  Amanda and Sampson,  Margaret and Shields,  Tracy and Stevens,  Adrienne and Sutton,  Anthea and Weinfurter,  Elizabeth and Wright,  Kath and Young,  Sarah},
  year = {2021},
  month = jan 
}

@article{Abouelela2023,
  title = {Exploring the Potentials of Open-Source Big Data and Machine Learning in Shared Mobility Fleet Utilization Prediction},
  volume = {5},
  ISSN = {2948-1368},
  DOI = {10.1007/s42421-023-00068-9},
  number = {2},
  journal = {Data Science for Transportation},
  publisher = {Springer Science and Business Media LLC},
  author = {Abouelela,  Mohamed and Lyu,  Cheng and Antoniou,  Constantinos},
  year = {2023},
  month = apr 
}

@article{Shi2023,
  title = {Short-Term Demand Prediction of Shared Bikes Based on LSTM Network},
  volume = {12},
  ISSN = {2079-9292},
  DOI = {10.3390/electronics12061381},
  number = {6},
  journal = {Electronics},
  publisher = {MDPI AG},
  author = {Shi,  Yi and Zhang,  Liumei and Lu,  Shengnan and Liu,  Qiao},
  year = {2023},
  month = mar,
  pages = {1381}
}

@article{Golpayegani2022,
  title = {Intelligent Shared Mobility Systems: A Survey on Whole System Design Requirements,  Challenges and Future Direction},
  volume = {10},
  ISSN = {2169-3536},
  DOI = {10.1109/access.2022.3162848},
  journal = {IEEE Access},
  publisher = {Institute of Electrical and Electronics Engineers (IEEE)},
  author = {Golpayegani,  Fatemeh and Gueriau,  Maxime and Laharotte,  Pierre-Antoine and Ghanadbashi,  Saeedeh and Guo,  Jiaying and Geraghty,  Jack and Wang,  Shen},
  year = {2022},
  pages = {35302–35320}
}

@article{Albuquerque2021,
  title = {Machine Learning Approaches to Bike-Sharing Systems: A Systematic Literature Review},
  volume = {10},
  ISSN = {2220-9964},
  DOI = {10.3390/ijgi10020062},
  number = {2},
  journal = {ISPRS International Journal of Geo-Information},
  publisher = {MDPI AG},
  author = {Albuquerque,  Vitória and Sales Dias,  Miguel and Bacao,  Fernando},
  year = {2021},
  month = feb,
  pages = {62}
}

@inproceedings{Tsiligkaridis2022,
  title = {Context-Aware Destination and Time-To-Destination Prediction Using Machine learning},
  DOI = {10.1109/isc255366.2022.9922593},
  booktitle = {2022 IEEE International Smart Cities Conference (ISC2)},
  publisher = {IEEE},
  author = {Tsiligkaridis,  Athanasios and Zhang,  Jing and Paschalidis,  Ioannis Ch. and Taguchi,  Hiroshi and Sakajo,  Satoko and Nikovski,  Daniel},
  year = {2022},
  month = sep 
}

@article{Zhao2022,
  title = {MDLF: A Multi-View-Based Deep Learning Framework for Individual Trip Destination Prediction in Public Transportation Systems},
  volume = {23},
  ISSN = {1558-0016},
  DOI = {10.1109/tits.2021.3123342},
  number = {8},
  journal = {IEEE Transactions on Intelligent Transportation Systems},
  publisher = {Institute of Electrical and Electronics Engineers (IEEE)},
  author = {Zhao,  Juanjuan and Zhang,  Liutao and Ye,  Jiexia and Xu,  Chengzhong},
  year = {2022},
  month = aug,
  pages = {13316–13329}
}

@article{Yang2024,
  title = {AI-driven anonymization: Protecting personal data privacy while leveraging machine learning},
  volume = {67},
  ISSN = {2755-273X},
  DOI = {10.54254/2755-2721/67/2024ma0053},
  number = {1},
  journal = {Applied and Computational Engineering},
  publisher = {EWA Publishing},
  author = {Yang,  Le and Tian,  Miao and Xin,  Duan and Cheng,  Qishuo and Zheng,  Jiajian},
  year = {2024},
  month = may,
  pages = {256–262}
}

@inproceedings{Himthani2020,
  title = {Big Data Privacy and Challenges for Machine Learning},
  DOI = {10.1109/i-smac49090.2020.9243527},
  booktitle = {2020 Fourth International Conference on I-SMAC (IoT in Social,  Mobile,  Analytics and Cloud) (I-SMAC)},
  publisher = {IEEE},
  author = {Himthani,  Puneet and Dubey,  Ghanshyam Prasad and Sharma,  Brij Mohan and Taneja,  Ankur},
  year = {2020},
  month = oct 
}

@article{Ullah2021,
  title = {A comparative performance of machine learning algorithm to predict electric vehicles energy consumption: A path towards sustainability},
  volume = {33},
  ISSN = {2048-4070},
  DOI = {10.1177/0958305x211044998},
  number = {8},
  journal = {Energy \& Environment},
  publisher = {SAGE Publications},
  author = {Ullah,  Irfan and Liu,  Kai and Yamamoto,  Toshiyuki and Al Mamlook,  Rabia Emhamed and Jamal,  Arshad},
  year = {2021},
  month = oct,
  pages = {1583–1612}
}

@article{Nabi2023,
  title = {Parametric analysis and prediction of energy consumption of electric vehicles using machine learning},
  volume = {72},
  ISSN = {2352-152X},
  DOI = {10.1016/j.est.2023.108226},
  journal = {Journal of Energy Storage},
  publisher = {Elsevier BV},
  author = {Nabi,  Md. Nurun and Ray,  Biplob and Rashid,  Fazlur and Al Hussam,  Wisam and Muyeen,  S.M.},
  year = {2023},
  month = nov,
  pages = {108226}
}

@article{Pokharel2021,
  title = {Improved Prediction of Total Energy Consumption and Feature Analysis in Electric Vehicles Using Machine Learning and Shapley Additive Explanations Method},
  volume = {12},
  ISSN = {2032-6653},
  DOI = {10.3390/wevj12030094},
  number = {3},
  journal = {World Electric Vehicle Journal},
  publisher = {MDPI AG},
  author = {Pokharel,  Sugam and Sah,  Pradip and Ganta,  Deepak},
  year = {2021},
  month = jun,
  pages = {94}
}

@article{Yan2023_1,
  title = {Privacy-Aware Energy Consumption Modeling of Connected Battery Electric Vehicles using Federated Learning},
  ISSN = {2372-2088},
  DOI = {10.1109/tte.2023.3343106},
  journal = {IEEE Transactions on Transportation Electrification},
  publisher = {Institute of Electrical and Electronics Engineers (IEEE)},
  author = {Yan,  Sen and Fang,  Hongyuan and Li,  Ji and Ward,  Tomas and O’Connor,  Noel and Liu,  Mingming},
  year = {2023},
  pages = {1–1}
}

@inproceedings{Yan2023_2,
  title = {Breathing Green: Maximising Health and Environmental Benefits for Active Transportation Users Leveraging Large Scale Air Quality Data},
  DOI = {10.1109/itsc57777.2023.10421940},
  booktitle = {2023 IEEE 26th International Conference on Intelligent Transportation Systems (ITSC)},
  publisher = {IEEE},
  author = {Yan,  Sen and Zhu,  Shaoshu and Fernandez,  Jaime B. and Sánchez,  Eric Arazo and Gu,  Yingqi and O’Connor,  Noel E. and O’Connor,  David and Liu,  Mingming},
  year = {2023},
  month = sep 
}

@inproceedings{Ding2024,
  title = {Data-driven Energy Consumption Modelling for Electric Micromobility using an Open Dataset},
  DOI = {10.1109/itec60657.2024.10599070},
  booktitle = {2024 IEEE Transportation Electrification Conference and Expo (ITEC)},
  publisher = {IEEE},
  author = {Ding,  Yue and Yan,  Sen and Shah,  Maqsood Hussain and Fang,  Hongyuan and Li,  Ji and Liu,  Mingming},
  year = {2024},
  month = jun 
}

@misc{Shah2024,
  doi = {10.48550/ARXIV.2403.07964},
  author = {Shah,  Maqsood Hussain and Ding,  Yue and Zhu,  Shaoshu and Gu,  Yingqi and Liu,  Mingming},
  title = {Optimal Design and Implementation of an Open-source Emulation Platform for User-Centric Shared E-mobility Services},
  publisher = {arXiv},
  year = {2024},
  copyright = {Creative Commons Attribution 4.0 International}
}

@article{Aboeleneen2024,
  title = {Reinforcement Learning-Based E-Scooter Energy Minimization Using Optimized Speed-Route Selection},
  volume = {12},
  ISSN = {2169-3536},
  DOI = {10.1109/access.2024.3395286},
  journal = {IEEE Access},
  publisher = {Institute of Electrical and Electronics Engineers (IEEE)},
  author = {Aboeleneen,  Karim and Zorba,  Nizar and Massoud,  Ahmed M.},
  year = {2024},
  pages = {66167–66184}
}

@article{Lv2022,
  title = {Dynamic Programming BN Structure Learning Algorithm Integrating Double Constraints under Small Sample Condition},
  volume = {24},
  ISSN = {1099-4300},
  DOI = {10.3390/e24101354},
  number = {10},
  journal = {Entropy},
  publisher = {MDPI AG},
  author = {Lv,  Zhigang and Chen,  Yiwei and Di,  Ruohai and Wang,  Hongxi and Sun,  Xiaojing and He,  Chuchao and Li,  Xiaoyan},
  year = {2022},
  month = sep,
  pages = {1354}
}

@article{Hou2023,
  title = {SoC-VRP: A Deep-Reinforcement-Learning-Based Vehicle Route Planning Mechanism for Service-Oriented Cooperative ITS},
  volume = {12},
  ISSN = {2079-9292},
  DOI = {10.3390/electronics12204191},
  number = {20},
  journal = {Electronics},
  publisher = {MDPI AG},
  author = {Hou,  Boyuan and Zhang,  Kailong and Gong,  Zu and Li,  Qiugang and Zhou,  Junle and Zhang,  Jiahao and de La Fortelle,  Arnaud},
  year = {2023},
  month = oct,
  pages = {4191}
}

@article{Ullah2021_1,
  title = {Electric vehicle energy consumption prediction using stacked generalization: an ensemble learning approach},
  volume = {18},
  ISSN = {1543-5083},
  DOI = {10.1080/15435075.2021.1881902},
  number = {9},
  journal = {International Journal of Green Energy},
  publisher = {Informa UK Limited},
  author = {Ullah,  Irfan and Liu,  Kai and Yamamoto,  Toshiyuki and Zahid,  Muhammad and Jamal,  Arshad},
  year = {2021},
  month = feb,
  pages = {896–909}
}

@article{Abdelaty2021,
  title = {Machine learning prediction models for battery-electric bus energy consumption in transit},
  volume = {96},
  ISSN = {1361-9209},
  DOI = {10.1016/j.trd.2021.102868},
  journal = {Transportation Research Part D: Transport and Environment},
  publisher = {Elsevier BV},
  author = {Abdelaty,  Hatem and Al-Obaidi,  Abdullah and Mohamed,  Moataz and Farag,  Hany E.Z.},
  year = {2021},
  month = jul,
  pages = {102868}
}

@article{Modi2020,
  title = {Estimation of energy consumption of electric vehicles using Deep Convolutional Neural Network to reduce driver’s range anxiety},
  volume = {98},
  ISSN = {0019-0578},
  DOI = {10.1016/j.isatra.2019.08.055},
  journal = {ISA Transactions},
  publisher = {Elsevier BV},
  author = {Modi,  Shatrughan and Bhattacharya,  Jhilik and Basak,  Prasenjit},
  year = {2020},
  month = mar,
  pages = {454–470}
}

@book{Montgomery2021,
  title={Introduction to linear regression analysis},
  author={Montgomery, Douglas C and Peck, Elizabeth A and Vining, G Geoffrey},
  year={2021},
  publisher={John Wiley \& Sons}
}

@article{Li2023,
  title = {Exploring Time Series Models for Wind Speed Forecasting: A Comparative Analysis},
  volume = {16},
  ISSN = {1996-1073},
  DOI = {10.3390/en16237785},
  number = {23},
  journal = {Energies},
  publisher = {MDPI AG},
  author = {Li,  Xiangqian and Li,  Keke and Shen,  Siqi and Tian,  Yaxin},
  year = {2023},
  month = nov,
  pages = {7785}
}

@inproceedings{Jani2023,
  title = {SAM based Computational Analysis of Root Mean Square Error for SPES Using LR,  RF \& ARIMA Machine Learning Models},
  DOI = {10.1109/icefeet59656.2023.10452190},
  booktitle = {2023 3rd International Conference on Emerging Frontiers in Electrical and Electronic Technologies (ICEFEET)},
  publisher = {IEEE},
  author = {Jani,  Rudresh and Patel,  Palak and Joshi,  Siddharth and Vyas,  Soham},
  year = {2023},
  month = dec 
}

@inproceedings{Bittencourt2024,
  title = {Towards Lightweight Fire Detection at the Extreme Edge based on Decision Trees},
  DOI = {10.1109/melecon56669.2024.10608598},
  booktitle = {2024 IEEE 22nd Mediterranean Electrotechnical Conference (MELECON)},
  publisher = {IEEE},
  author = {Bittencourt,  João Carlos N. and Costa,  Daniel G. and Portugal,  Paulo and Vasques,  Francisco},
  year = {2024},
  month = jun 
}

@inproceedings{Liu2019_1,
  title = {Defect Rating for Different Production Models of Line Protection Devices Based on Decision Trees},
  DOI = {10.1109/ei247390.2019.9062248},
  booktitle = {2019 IEEE 3rd Conference on Energy Internet and Energy System Integration (EI2)},
  publisher = {IEEE},
  author = {Liu,  Yimin and Yang,  Xinping and Niu,  Siqing and Zheng,  Shaoming and Chen,  Qian and Xue,  Ancheng},
  year = {2019},
  month = nov 
}

@inproceedings{Moumen2020,
  title = {Data mining techniques for employability: Systematic literature review},
  DOI = {10.1109/icecocs50124.2020.9314555},
  booktitle = {2020 IEEE 2nd International Conference on Electronics,  Control,  Optimization and Computer Science (ICECOCS)},
  publisher = {IEEE},
  author = {Moumen,  Aniss and Bouchama,  El Houcine and El Bouzekri El Idirissi,  Younes},
  year = {2020},
  month = dec 
}

@article{Faizi2024,
  title = {Improved Segmentation Model for Melanoma Lesion Detection Using Normalized Cross-Correlation-Based k-Means Clustering},
  volume = {12},
  ISSN = {2169-3536},
  DOI = {10.1109/access.2024.3360223},
  journal = {IEEE Access},
  publisher = {Institute of Electrical and Electronics Engineers (IEEE)},
  author = {Faizi,  Muhammad Imran and Adnan,  Syed Muhammad},
  year = {2024},
  pages = {20753–20766}
}

@article{Borisov2024,
  title = {Deep Neural Networks and Tabular Data: A Survey},
  volume = {35},
  ISSN = {2162-2388},
  DOI = {10.1109/tnnls.2022.3229161},
  number = {6},
  journal = {IEEE Transactions on Neural Networks and Learning Systems},
  publisher = {Institute of Electrical and Electronics Engineers (IEEE)},
  author = {Borisov,  Vadim and Leemann,  Tobias and Seßler,  Kathrin and Haug,  Johannes and Pawelczyk,  Martin and Kasneci,  Gjergji},
  year = {2024},
  month = jun,
  pages = {7499–7519}
}

@article{Bratsas2019,
  title = {A Comparison of Machine Learning Methods for the Prediction of Traffic Speed in Urban Places},
  volume = {12},
  ISSN = {2071-1050},
  DOI = {10.3390/su12010142},
  number = {1},
  journal = {Sustainability},
  publisher = {MDPI AG},
  author = {Bratsas,  Charalampos and Koupidis,  Kleanthis and Salanova,  Josep-Maria and Giannakopoulos,  Konstantinos and Kaloudis,  Aristeidis and Aifadopoulou,  Georgia},
  year = {2019},
  month = dec,
  pages = {142}
}

@article{Sehovac2020,
  title = {Deep Learning for Load Forecasting: Sequence to Sequence Recurrent Neural Networks With Attention},
  volume = {8},
  ISSN = {2169-3536},
  DOI = {10.1109/access.2020.2975738},
  journal = {IEEE Access},
  publisher = {Institute of Electrical and Electronics Engineers (IEEE)},
  author = {Sehovac,  Ljubisa and Grolinger,  Katarina},
  year = {2020},
  pages = {36411–36426}
}

@article{Hewamalage2021,
  title = {Recurrent Neural Networks for Time Series Forecasting: Current status and future directions},
  volume = {37},
  ISSN = {0169-2070},
  DOI = {10.1016/j.ijforecast.2020.06.008},
  number = {1},
  journal = {International Journal of Forecasting},
  publisher = {Elsevier BV},
  author = {Hewamalage,  Hansika and Bergmeir,  Christoph and Bandara,  Kasun},
  year = {2021},
  month = jan,
  pages = {388–427}
}

@inproceedings{Chen2021,
  title = {A Comparative Study of Using Spatial-Temporal Graph Convolutional Networks for Predicting Availability in Bike Sharing Schemes},
  DOI = {10.1109/itsc48978.2021.9564831},
  booktitle = {2021 IEEE International Intelligent Transportation Systems Conference (ITSC)},
  publisher = {IEEE},
  author = {Chen,  Zhengyong and Wu,  Hongde and O’Connor,  Noel E. and Liu,  Mingming},
  year = {2021},
  month = sep 
}

@inproceedings{Wu2022,
  title = {Lane-GNN: Integrating GNN for Predicting Drivers’ Lane Change Intention},
  DOI = {10.1109/itsc55140.2022.9922139},
  booktitle = {2022 IEEE 25th International Conference on Intelligent Transportation Systems (ITSC)},
  publisher = {IEEE},
  author = {Wu,  Hongde and Liu,  Mingming},
  year = {2022},
  month = oct 
}

@article{Sabri2024,
  title = {Detection of Micromobility Vehicles in Urban Traffic Videos},
  DOI = {10.21428/d82e957c.abc3243f},
  journal = {Proceedings of the 21st Conference on Robots and Vision},
  publisher = {PubPub},
  author = {Sabri,  Khalil and Djilali,  Célia and Bilodeau,  Guillaume-Alexandre and Saunier,  Nicolas and Bouachir,  Wassim},
  year = {2024},
  month = may 
}

@misc{Chen2024,
  doi = {10.5281/ZENODO.10578641},
  author = {Chen,  Dong},
  title = {Object Detection for E-scooters},
  publisher = {Zenodo},
  year = {2024},
  copyright = {Creative Commons Attribution 4.0 International}
}

@article{Jang2023,
  title = {An AI Safety Monitoring System for Electric Scooters Based on the Number of Riders and Road Types},
  volume = {23},
  ISSN = {1424-8220},
  DOI = {10.3390/s23229181},
  number = {22},
  journal = {Sensors},
  publisher = {MDPI AG},
  author = {Jang,  Woo-Jin and Kim,  Dong-Hyun and Lim,  Si-Hyung},
  year = {2023},
  month = nov,
  pages = {9181}
}

@article{Tamagusko2023,
  title = {Data-Driven Approach for Urban Micromobility Enhancement through Safety Mapping and Intelligent Route Planning},
  volume = {6},
  ISSN = {2624-6511},
  DOI = {10.3390/smartcities6040094},
  number = {4},
  journal = {Smart Cities},
  publisher = {MDPI AG},
  author = {Tamagusko,  Tiago and Gomes Correia,  Matheus and Rita,  Luís and Bostan,  Tudor-Codrin and Peliteiro,  Miguel and Martins,  Rodrigo and Santos,  Luísa and Ferreira,  Adelino},
  year = {2023},
  month = aug,
  pages = {2035–2056}
}

@inbook{Nguyen2021,
  title = {Electric Scooter and Its Rider Detection Framework Based on Deep Learning for Supporting Scooter-Related Injury Emergency Services},
  ISBN = {9783030720735},
  ISSN = {1865-0937},
  DOI = {10.1007/978-3-030-72073-5_18},
  booktitle = {Geometry and Vision},
  publisher = {Springer International Publishing},
  author = {Nguyen,  Hoa and Nguyen,  Minh and Sun,  Qian},
  year = {2021},
  pages = {233–246}
}

@inproceedings{Khalil2021,
  title = {Traffic Study of Shared Micromobility Services by Transportation Simulation},
  DOI = {10.1109/bigdata52589.2021.9671455},
  booktitle = {2021 IEEE International Conference on Big Data (Big Data)},
  publisher = {IEEE},
  author = {Khalil,  Jalal and Yan,  Da and Guo,  Guimu and Sami,  Mirza Tanzim and Roy,  Joy Bhadhan and Sisiopiku,  Virginia P.},
  year = {2021},
  month = dec 
}

@article{Geiger2013,
  title = {Vision meets robotics: The KITTI dataset},
  volume = {32},
  ISSN = {1741-3176},
  DOI = {10.1177/0278364913491297},
  number = {11},
  journal = {The International Journal of Robotics Research},
  publisher = {SAGE Publications},
  author = {Geiger,  A and Lenz,  P and Stiller,  C and Urtasun,  R},
  year = {2013},
  month = aug,
  pages = {1231–1237}
}

@inproceedings{Cheng2019,
  title = {DET: A High-Resolution DVS Dataset for Lane Extraction},
  DOI = {10.1109/cvprw.2019.00210},
  booktitle = {2019 IEEE/CVF Conference on Computer Vision and Pattern Recognition Workshops (CVPRW)},
  publisher = {IEEE},
  author = {Cheng,  Wensheng and Luo,  Hao and Yang,  Wen and Yu,  Lei and Chen,  Shoushun and Li,  Wei},
  year = {2019},
  month = jun 
}

@inproceedings{Chang2019,
  title = {Argoverse: 3D Tracking and Forecasting With Rich Maps},
  DOI = {10.1109/cvpr.2019.00895},
  booktitle = {2019 IEEE/CVF Conference on Computer Vision and Pattern Recognition (CVPR)},
  publisher = {IEEE},
  author = {Chang,  Ming-Fang and Ramanan,  Deva and Hays,  James and Lambert,  John and Sangkloy,  Patsorn and Singh,  Jagjeet and Bak,  Slawomir and Hartnett,  Andrew and Wang,  De and Carr,  Peter and Lucey,  Simon},
  year = {2019},
  month = jun 
}

@inproceedings{Xiao2021,
  title = {PandaSet: Advanced Sensor Suite Dataset for Autonomous Driving},
  DOI = {10.1109/itsc48978.2021.9565009},
  booktitle = {2021 IEEE International Intelligent Transportation Systems Conference (ITSC)},
  publisher = {IEEE},
  author = {Xiao,  Pengchuan and Shao,  Zhenlei and Hao,  Steven and Zhang,  Zishuo and Chai,  Xiaolin and Jiao,  Judy and Li,  Zesong and Wu,  Jian and Sun,  Kai and Jiang,  Kun and Wang,  Yunlong and Yang,  Diange},
  year = {2021},
  month = sep 
}

@inproceedings{Sun2020,
  title = {Scalability in Perception for Autonomous Driving: Waymo Open Dataset},
  DOI = {10.1109/cvpr42600.2020.00252},
  booktitle = {2020 IEEE/CVF Conference on Computer Vision and Pattern Recognition (CVPR)},
  publisher = {IEEE},
  author = {Sun,  Pei and Kretzschmar,  Henrik and Dotiwalla,  Xerxes and Chouard,  Aurelien and Patnaik,  Vijaysai and Tsui,  Paul and Guo,  James and Zhou,  Yin and Chai,  Yuning and Caine,  Benjamin and Vasudevan,  Vijay and Han,  Wei and Ngiam,  Jiquan and Zhao,  Hang and Timofeev,  Aleksei and Ettinger,  Scott and Krivokon,  Maxim and Gao,  Amy and Joshi,  Aditya and Zhang,  Yu and Shlens,  Jonathon and Chen,  Zhifeng and Anguelov,  Dragomir},
  year = {2020},
  month = jun 
}

@inproceedings{Caesar2020,
  title = {nuScenes: A Multimodal Dataset for Autonomous Driving},
  DOI = {10.1109/cvpr42600.2020.01164},
  booktitle = {2020 IEEE/CVF Conference on Computer Vision and Pattern Recognition (CVPR)},
  publisher = {IEEE},
  author = {Caesar,  Holger and Bankiti,  Varun and Lang,  Alex H. and Vora,  Sourabh and Liong,  Venice Erin and Xu,  Qiang and Krishnan,  Anush and Pan,  Yu and Baldan,  Giancarlo and Beijbom,  Oscar},
  year = {2020},
  month = jun 
}

@article{Jaber2023,
  title = {The Preferences of Shared Micro-Mobility Users in Urban Areas},
  volume = {11},
  ISSN = {2169-3536},
  DOI = {10.1109/access.2023.3297083},
  journal = {IEEE Access},
  publisher = {Institute of Electrical and Electronics Engineers (IEEE)},
  author = {Jaber,  Ahmed and Hamadneh,  Jamil and Csonka,  Bálint},
  year = {2023},
  pages = {74458–74472}
}

@inproceedings{Yan2022,
  title = {Parking Behaviour Analysis of Shared E-Bike Users Based on a Real-World Dataset - A Case Study in Dublin,  Ireland},
  DOI = {10.1109/vtc2022-spring54318.2022.9860871},
  booktitle = {2022 IEEE 95th Vehicular Technology Conference: (VTC2022-Spring)},
  publisher = {IEEE},
  author = {Yan,  Sen and Liu,  Mingming and O’Connor,  Noel E.},
  year = {2022},
  month = jun 
}

@article{VegaGonzalo2024,
  title = {The role of shared mobility in reducing perceived private car dependency},
  volume = {126},
  ISSN = {1361-9209},
  DOI = {10.1016/j.trd.2023.104023},
  journal = {Transportation Research Part D: Transport and Environment},
  publisher = {Elsevier BV},
  author = {Vega-Gonzalo,  María and Gomez,  Juan and Christidis,  Panayotis and Manuel Vassallo,  José},
  year = {2024},
  month = jan,
  pages = {104023}
}

@article{Liu2022,
  title = {Measuring the impacts of dockless micro-mobility services on public transit accessibility},
  volume = {98},
  ISSN = {0198-9715},
  DOI = {10.1016/j.compenvurbsys.2022.101885},
  journal = {Computers,  Environment and Urban Systems},
  publisher = {Elsevier BV},
  author = {Liu,  Luyu and Miller,  Harvey J.},
  year = {2022},
  month = dec,
  pages = {101885}
}

@article{Almukhalfi2024,
  title = {Traffic management approaches using machine learning and deep learning techniques: A survey},
  volume = {133},
  ISSN = {0952-1976},
  DOI = {10.1016/j.engappai.2024.108147},
  journal = {Engineering Applications of Artificial Intelligence},
  publisher = {Elsevier BV},
  author = {Almukhalfi,  Hanan and Noor,  Ayman and Noor,  Talal H.},
  year = {2024},
  month = jul,
  pages = {108147}
}

@article{Modi2021,
  title = {A comprehensive review on intelligent traffic management using machine learning algorithms},
  volume = {7},
  ISSN = {2364-4184},
  DOI = {10.1007/s41062-021-00718-3},
  number = {1},
  journal = {Innovative Infrastructure Solutions},
  publisher = {Springer Science and Business Media LLC},
  author = {Modi,  Yash and Teli,  Ridham and Mehta,  Akshat and Shah,  Konark and Shah,  Manan},
  year = {2021},
  month = dec 
}

@article{Boukerche2020,
  title = {Machine Learning-based traffic prediction models for Intelligent Transportation Systems},
  volume = {181},
  ISSN = {1389-1286},
  DOI = {10.1016/j.comnet.2020.107530},
  journal = {Computer Networks},
  publisher = {Elsevier BV},
  author = {Boukerche,  Azzedine and Wang,  Jiahao},
  year = {2020},
  month = nov,
  pages = {107530}
}

@inproceedings{Sandler2018,
  title = {MobileNetV2: Inverted Residuals and Linear Bottlenecks},
  DOI = {10.1109/cvpr.2018.00474},
  booktitle = {2018 IEEE/CVF Conference on Computer Vision and Pattern Recognition},
  publisher = {IEEE},
  author = {Sandler,  Mark and Howard,  Andrew and Zhu,  Menglong and Zhmoginov,  Andrey and Chen,  Liang-Chieh},
  year = {2018},
  month = jun 
}

@inproceedings{zhang2018shufflenet,
  title={Shufflenet: An extremely efficient convolutional neural network for mobile devices},
  author={Zhang, Xiangyu and Zhou, Xinyu and Lin, Mengxiao and Sun, Jian},
  booktitle={Proceedings of the IEEE conference on computer vision and pattern recognition},
  pages={6848--6856},
  year={2018}
}

@article{iandola2016squeezenet,
  title={SqueezeNet: AlexNet-level accuracy with 50x fewer parameters and< 0.5 MB model size},
  author={Iandola, Forrest N and Han, Song and Moskewicz, Matthew W and Ashraf, Khalid and Dally, William J and Keutzer, Kurt},
  journal={arXiv preprint arXiv:1602.07360},
  year={2016}
}

@article{He2021,
  title = {AutoML: A survey of the state-of-the-art},
  volume = {212},
  ISSN = {0950-7051},
  DOI = {10.1016/j.knosys.2020.106622},
  journal = {Knowledge-Based Systems},
  publisher = {Elsevier BV},
  author = {He,  Xin and Zhao,  Kaiyong and Chu,  Xiaowen},
  year = {2021},
  month = jan,
  pages = {106622}
}

@inproceedings{Amershi2019,
  title = {Software Engineering for Machine Learning: A Case Study},
  DOI = {10.1109/icse-seip.2019.00042},
  booktitle = {2019 IEEE/ACM 41st International Conference on Software Engineering: Software Engineering in Practice (ICSE-SEIP)},
  publisher = {IEEE},
  author = {Amershi,  Saleema and Begel,  Andrew and Bird,  Christian and DeLine,  Robert and Gall,  Harald and Kamar,  Ece and Nagappan,  Nachiappan and Nushi,  Besmira and Zimmermann,  Thomas},
  year = {2019},
  month = may 
}

@article{Lopes2014,
  title = {GIS-Based Analytical Tools for Transport Planning:  Spatial Regression Models for Transportation Demand Forecast},
  volume = {3},
  ISSN = {2220-9964},
  DOI = {10.3390/ijgi3020565},
  number = {2},
  journal = {ISPRS International Journal of Geo-Information},
  publisher = {MDPI AG},
  author = {Lopes,  Simone and Brondino,  Nair and Rodrigues da Silva,  Ant\^onio},
  year = {2014},
  month = apr,
  pages = {565–583}
}

@article{Ceha1997,
  title = {Prediction of future origin-destination matrix of air passengers by fratar and gravity models},
  volume = {33},
  ISSN = {0360-8352},
  DOI = {10.1016/s0360-8352(97)00263-5},
  number = {3–4},
  journal = {Computers \& Industrial Engineering},
  publisher = {Elsevier BV},
  author = {Ceha,  Rakhmat and Ohta,  Hiroshi},
  year = {1997},
  month = dec,
  pages = {845–848}
}

@article{Liu2017,
  title = {An application of network analysis on tourist attractions: The case of Xinjiang,  China},
  volume = {58},
  ISSN = {0261-5177},
  DOI = {10.1016/j.tourman.2016.10.009},
  journal = {Tourism Management},
  publisher = {Elsevier BV},
  author = {Liu,  Bing and Huang,  Songshan (Sam) and Fu,  Hui},
  year = {2017},
  month = feb,
  pages = {132–141}
}

@misc{Dosovitskiy2020,
  doi = {10.48550/ARXIV.2010.11929},
  author = {Dosovitskiy,  Alexey and Beyer,  Lucas and Kolesnikov,  Alexander and Weissenborn,  Dirk and Zhai,  Xiaohua and Unterthiner,  Thomas and Dehghani,  Mostafa and Minderer,  Matthias and Heigold,  Georg and Gelly,  Sylvain and Uszkoreit,  Jakob and Houlsby,  Neil},
  title = {An Image is Worth 16x16 Words: Transformers for Image Recognition at Scale},
  publisher = {arXiv},
  year = {2020}
}

@article{Guruji2016,
  title = {Time-efficient A* Algorithm for Robot Path Planning},
  volume = {23},
  ISSN = {2212-0173},
  DOI = {10.1016/j.protcy.2016.03.010},
  journal = {Procedia Technology},
  publisher = {Elsevier BV},
  author = {Guruji,  Akshay Kumar and Agarwal,  Himansh and Parsediya,  D.K.},
  year = {2016},
  pages = {144–149}
}

@inproceedings{Fan2010,
  title = {Improvement of Dijkstra’s algorithm and its application in route planning},
  DOI = {10.1109/fskd.2010.5569452},
  booktitle = {2010 Seventh International Conference on Fuzzy Systems and Knowledge Discovery},
  publisher = {IEEE},
  author = {Fan,  DongKai and Shi,  Ping},
  year = {2010},
  month = aug,
  pages = {1901–1904}
}

@inbook{Chen2022,
  title = {PersFormer: 3D Lane Detection via Perspective Transformer and the OpenLane Benchmark},
  ISBN = {9783031198397},
  ISSN = {1611-3349},
  DOI = {10.1007/978-3-031-19839-7_32},
  booktitle = {Computer Vision – ECCV 2022},
  publisher = {Springer Nature Switzerland},
  author = {Chen,  Li and Sima,  Chonghao and Li,  Yang and Zheng,  Zehan and Xu,  Jiajie and Geng,  Xiangwei and Li,  Hongyang and He,  Conghui and Shi,  Jianping and Qiao,  Yu and Yan,  Junchi},
  year = {2022},
  pages = {550–567}
}

@inproceedings{Peng2023,
  title = {BEVSegFormer: Bird’s Eye View Semantic Segmentation From Arbitrary Camera Rigs},
  DOI = {10.1109/wacv56688.2023.00588},
  booktitle = {2023 IEEE/CVF Winter Conference on Applications of Computer Vision (WACV)},
  publisher = {IEEE},
  author = {Peng,  Lang and Chen,  Zhirong and Fu,  Zhangjie and Liang,  Pengpeng and Cheng,  Erkang},
  year = {2023},
  month = jan,
  pages = {5924–5932}
}

@inproceedings{Liu2021,
  title = {End-to-end Lane Shape Prediction with Transformers},
  DOI = {10.1109/wacv48630.2021.00374},
  booktitle = {2021 IEEE Winter Conference on Applications of Computer Vision (WACV)},
  publisher = {IEEE},
  author = {Liu,  Ruijin and Yuan,  Zejian and Liu,  Tie and Xiong,  Zhiliang},
  year = {2021},
  month = jan 
}

@inproceedings{Bai2023,
  title = {CurveFormer: 3D Lane Detection by Curve Propagation with Curve Queries and Attention},
  DOI = {10.1109/icra48891.2023.10161160},
  booktitle = {2023 IEEE International Conference on Robotics and Automation (ICRA)},
  publisher = {IEEE},
  author = {Bai,  Yifeng and Chen,  Zhirong and Fu,  Zhangjie and Peng,  Lang and Liang,  Pengpeng and Cheng,  Erkang},
  year = {2023},
  month = may 
}

@article{Liu2024,
  title = {Lightweight Deep Learning for Resource-Constrained Environments: A Survey},
  volume = {56},
  ISSN = {1557-7341},
  DOI = {10.1145/3657282},
  number = {10},
  journal = {ACM Computing Surveys},
  publisher = {Association for Computing Machinery (ACM)},
  author = {Liu,  Hou-I and Galindo,  Marco and Xie,  Hongxia and Wong,  Lai-Kuan and Shuai,  Hong-Han and Li,  Yung-Hui and Cheng,  Wen-Huang},
  year = {2024},
  month = jun,
  pages = {1–42}
}

@techreport{ITF2020,
  title        = {Safe Micromobility},
  author       = {{International Transport Forum}},
  institution  = {International Transport Forum, OECD Publishing},
  type         = {International Transport Forum Policy Papers No. 85},
  year         = {2020},
  address      = {Paris},
  doi          = {10.1787/0b98fac1-en},
}

@article{Oeschger2020,
  title = {Micromobility and public transport integration: The current state of knowledge},
  volume = {89},
  ISSN = {1361-9209},
  DOI = {10.1016/j.trd.2020.102628},
  journal = {Transportation Research Part D: Transport and Environment},
  publisher = {Elsevier BV},
  author = {Oeschger,  Giulia and Carroll,  Páraic and Caulfield,  Brian},
  year = {2020},
  month = dec,
  pages = {102628}
}

@article{Yi2021,
  title = {An Automated Hyperparameter Search-Based Deep Learning Model for Highway Traffic Prediction},
  volume = {22},
  ISSN = {1558-0016},
  DOI = {10.1109/tits.2020.2987614},
  number = {9},
  journal = {IEEE Transactions on Intelligent Transportation Systems},
  publisher = {Institute of Electrical and Electronics Engineers (IEEE)},
  author = {Yi,  Hongsuk and Bui,  Khac-Hoai Nam},
  year = {2021},
  month = sep,
  pages = {5486–5495}
}

@article{Xu2023,
  title = {Multi-task supply-demand prediction and reliability analysis for docked bike-sharing systems via transformer-encoder-based neural processes},
  volume = {147},
  ISSN = {0968-090X},
  DOI = {10.1016/j.trc.2023.104015},
  journal = {Transportation Research Part C: Emerging Technologies},
  publisher = {Elsevier BV},
  author = {Xu,  Meng and Di,  Yining and Yang,  Hai and Chen,  Xiqun and Zhu,  Zheng},
  year = {2023},
  month = feb,
  pages = {104015}
}

@article{Xu2023_2,
  title = {Real-Time Forecasting of Dockless Scooter-Sharing Demand: A Spatio-Temporal Multi-Graph Transformer Approach},
  volume = {24},
  ISSN = {1558-0016},
  DOI = {10.1109/tits.2023.3239309},
  number = {8},
  journal = {IEEE Transactions on Intelligent Transportation Systems},
  publisher = {Institute of Electrical and Electronics Engineers (IEEE)},
  author = {Xu,  Yiming and Zhao,  Xilei and Zhang,  Xiaojian and Paliwal,  Mudit},
  year = {2023},
  month = aug,
  pages = {8507–8518}
}

@article{Rhmann2024,
  title = {Interpretable Bike-Sharing Activity Prediction with a Temporal Fusion Transformer to Unveil Influential Factors: A Case Study in Hamburg,  Germany},
  volume = {16},
  ISSN = {2071-1050},
  DOI = {10.3390/su16083230},
  number = {8},
  journal = {Sustainability},
  publisher = {MDPI AG},
  author = {R\"{u}hmann,  Sebastian and Leible,  Stephan and Lewandowski,  Tom},
  year = {2024},
  month = apr,
  pages = {3230}
}

@article{Roth2024,
  title={Resource-efficient neural networks for embedded systems},
  author={Roth, Wolfgang and Schindler, G{\"u}nther and Klein, Bernhard and Peharz, Robert and Tschiatschek, Sebastian and Fr{\"o}ning, Holger and Pernkopf, Franz and Ghahramani, Zoubin},
  journal={Journal of Machine Learning Research},
  volume={25},
  number={50},
  pages={1--51},
  year={2024}
}

@article{DeBock2022,
  title = {Road cycling safety scoring based on geospatial analysis,  computer vision and machine learning},
  volume = {82},
  ISSN = {1573-7721},
  DOI = {10.1007/s11042-022-13552-1},
  number = {6},
  journal = {Multimedia Tools and Applications},
  publisher = {Springer Science and Business Media LLC},
  author = {De Bock,  Jelle and Verstockt,  Steven},
  year = {2022},
  month = jul,
  pages = {8359–8380}
}

@article{Costa2024,
  title = {Unraveling the relation between cycling accidents and built environment typologies: Capturing spatial heterogeneity through a latent class discrete outcome model},
  volume = {200},
  ISSN = {0001-4575},
  DOI = {10.1016/j.aap.2024.107533},
  journal = {Accident Analysis \& Prevention},
  publisher = {Elsevier BV},
  author = {Costa,  Miguel and Lima Azevedo,  Carlos and Siebert,  Felix Wilhelm and Marques,  Manuel and Moura,  Filipe},
  year = {2024},
  month = jun,
  pages = {107533}
}

@article{Lyu2024,
  title = {On the Model Update Strategies for Supervised Learning in AIOps Solutions},
  volume = {33},
  ISSN = {1557-7392},
  DOI = {10.1145/3664599},
  number = {7},
  journal = {ACM Transactions on Software Engineering and Methodology},
  publisher = {Association for Computing Machinery (ACM)},
  author = {Lyu,  Yingzhe and Li,  Heng and Jiang,  Zhen Ming (Jack) and Hassan,  Ahmed E.},
  year = {2024},
  month = aug,
  pages = {1–38}
}

@article{Dozza2022,
  title = {A data-driven framework for the safe integration of micro-mobility into the transport system: Comparing bicycles and e-scooters in field trials},
  volume = {81},
  ISSN = {0022-4375},
  DOI = {10.1016/j.jsr.2022.01.007},
  journal = {Journal of Safety Research},
  publisher = {Elsevier BV},
  author = {Dozza,  Marco and Violin,  Alessio and Rasch,  Alexander},
  year = {2022},
  month = jun,
  pages = {67–77}
}

@article{Jiang2022,
  title = {Graph neural network for traffic forecasting: A survey},
  volume = {207},
  ISSN = {0957-4174},
  DOI = {10.1016/j.eswa.2022.117921},
  journal = {Expert Systems with Applications},
  publisher = {Elsevier BV},
  author = {Jiang,  Weiwei and Luo,  Jiayun},
  year = {2022},
  month = Nov,
  pages = {117921}
}

@inproceedings{Wu2019,
  series = {IJCAI-2019},
  title = {Graph WaveNet for Deep Spatial-Temporal Graph Modeling},
  DOI = {10.24963/ijcai.2019/264},
  booktitle = {Proceedings of the Twenty-Eighth International Joint Conference on Artificial Intelligence},
  publisher = {International Joint Conferences on Artificial Intelligence Organization},
  author = {Wu,  Zonghan and Pan,  Shirui and Long,  Guodong and Jiang,  Jing and Zhang,  Chengqi},
  year = {2019},
  month = Aug,
  pages = {1907–1913},
  collection = {IJCAI-2019}
}

@article{Belkessa2026,
  title = {Integrating subway checkout data for shared micromobility demand forecasting: A novel deep learning framework},
  volume = {187},
  ISSN = {0968-090X},
  DOI = {10.1016/j.trc.2026.105650},
  journal = {Transportation Research Part C: Emerging Technologies},
  publisher = {Elsevier BV},
  author = {Belkessa,  Linda and Ameli,  Mostafa and Ramezani,  Mohsen and Zargayouna,  Mahdi},
  year = {2026},
  month = Jun,
  pages = {105650}
}

@article{Zhu2025,
  title = {FedCGAT: A federated demand prediction method for shared bicycles based on contrastive graph attention mechanism},
  volume = {48},
  ISSN = {1875-8967},
  DOI = {10.3233/jifs-239708},
  number = {3},
  journal = {Journal of Intelligent \& Fuzzy Systems: Applications in Engineering and Technology},
  publisher = {SAGE Publications},
  author = {Zhu,  Yichang and Chen,  Leiming and Zhang,  Weishan and Zhang,  Baoyu and Dong,  Cihao and Qiu,  Jiakang},
  year = {2025},
  month = Feb,
  pages = {329–344}
}

@article{Yan2020,
  title = {Fairness-Aware Demand Prediction for New Mobility},
  volume = {34},
  ISSN = {2159-5399},
  DOI = {10.1609/aaai.v34i01.5458},
  number = {01},
  journal = {Proceedings of the AAAI Conference on Artificial Intelligence},
  publisher = {Association for the Advancement of Artificial Intelligence (AAAI)},
  author = {Yan,  An and Howe,  Bill},
  year = {2020},
  month = Apr,
  pages = {1079–1087}
}

@article{Yang2025,
  title = {Cross-Modality and Equity-Aware Graph Pooling Fusion: A Bike Mobility Prediction Study},
  volume = {11},
  ISSN = {2372-2096},
  DOI = {10.1109/tbdata.2024.3414280},
  number = {1},
  journal = {IEEE Transactions on Big Data},
  publisher = {Institute of Electrical and Electronics Engineers (IEEE)},
  author = {Yang,  Xi and He,  Suining and Shin,  Kang G. and Tabatabaie,  Mahan and Dai,  Jing},
  year = {2025},
  month = Feb,
  pages = {286–302}
}

@article{Singla2015,
  title = {Incentivizing Users for Balancing Bike Sharing Systems},
  volume = {29},
  ISSN = {2159-5399},
  DOI = {10.1609/aaai.v29i1.9251},
  number = {1},
  journal = {Proceedings of the AAAI Conference on Artificial Intelligence},
  publisher = {Association for the Advancement of Artificial Intelligence (AAAI)},
  author = {Singla,  Adish and Santoni,  Marco and Bartók,  Gábor and Mukerji,  Pratik and Meenen,  Moritz and Krause,  Andreas},
  year = {2015},
  month = Feb 
}

@article{Shi2024,
  title = {Multi-Platform dynamic game and operation of hybrid Bike-Sharing systems based on reinforcement learning},
  volume = {181},
  ISSN = {1366-5545},
  DOI = {10.1016/j.tre.2023.103374},
  journal = {Transportation Research Part E: Logistics and Transportation Review},
  publisher = {Elsevier BV},
  author = {Shi,  Ziyi and Xu,  Meng and Song,  Yancun and Zhu,  Zheng},
  year = {2024},
  month = Jan,
  pages = {103374}
}

@inproceedings{Chai2018,
  series = {SIGSPATIAL ’18},
  title = {Bike flow prediction with multi-graph convolutional networks},
  DOI = {10.1145/3274895.3274896},
  booktitle = {Proceedings of the 26th ACM SIGSPATIAL International Conference on Advances in Geographic Information Systems},
  publisher = {ACM},
  author = {Chai,  Di and Wang,  Leye and Yang,  Qiang},
  year = {2018},
  month = Nov,
  pages = {397–400},
  collection = {SIGSPATIAL ’18}
}

\vspace{-2\baselineskip}

\begin{IEEEbiography}[{
    \includegraphics[width=1in,height=1.33in,clip,keepaspectratio]{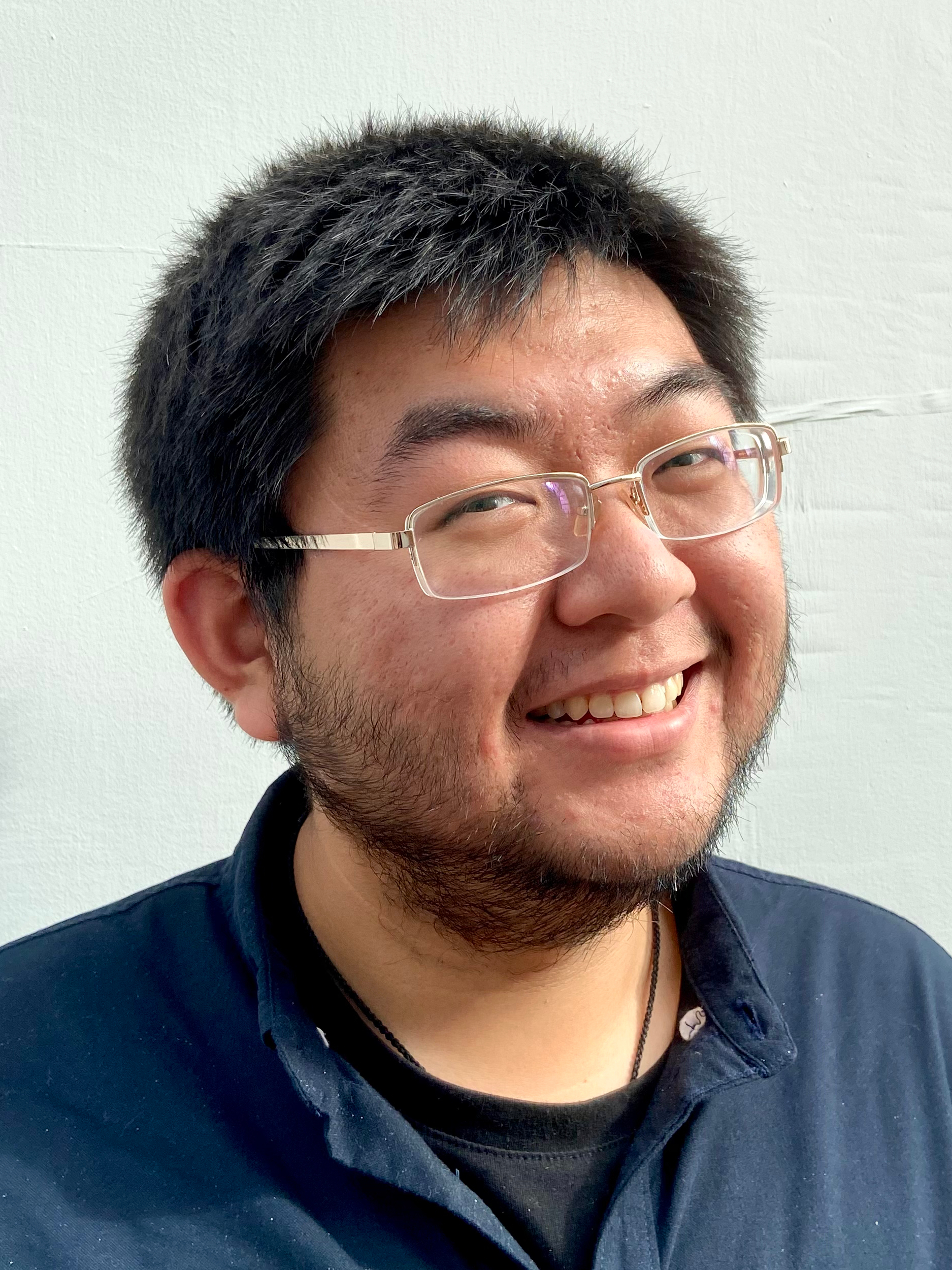}}]
    {Sen Yan (Graduate Student Member, IEEE)}
    received his B.Eng (1st Hons) in Internet of Things Engineering from Shandong University in 2018, his M.Sc (1st Hons) in Computer Science from Trinity College Dublin, the University of Dublin, in 2020, and his Ph.D. degree from the School of Electronic Engineering, Dublin City University, in 2025. His research focuses on data-driven approaches for intelligent transportation systems, especially shared micromobility systems. He is also interested in image processing and computer vision.
\end{IEEEbiography}

\vspace{-2\baselineskip}

\begin{IEEEbiography}[{
    \includegraphics[width=1in,height=1.33in,clip,keepaspectratio]{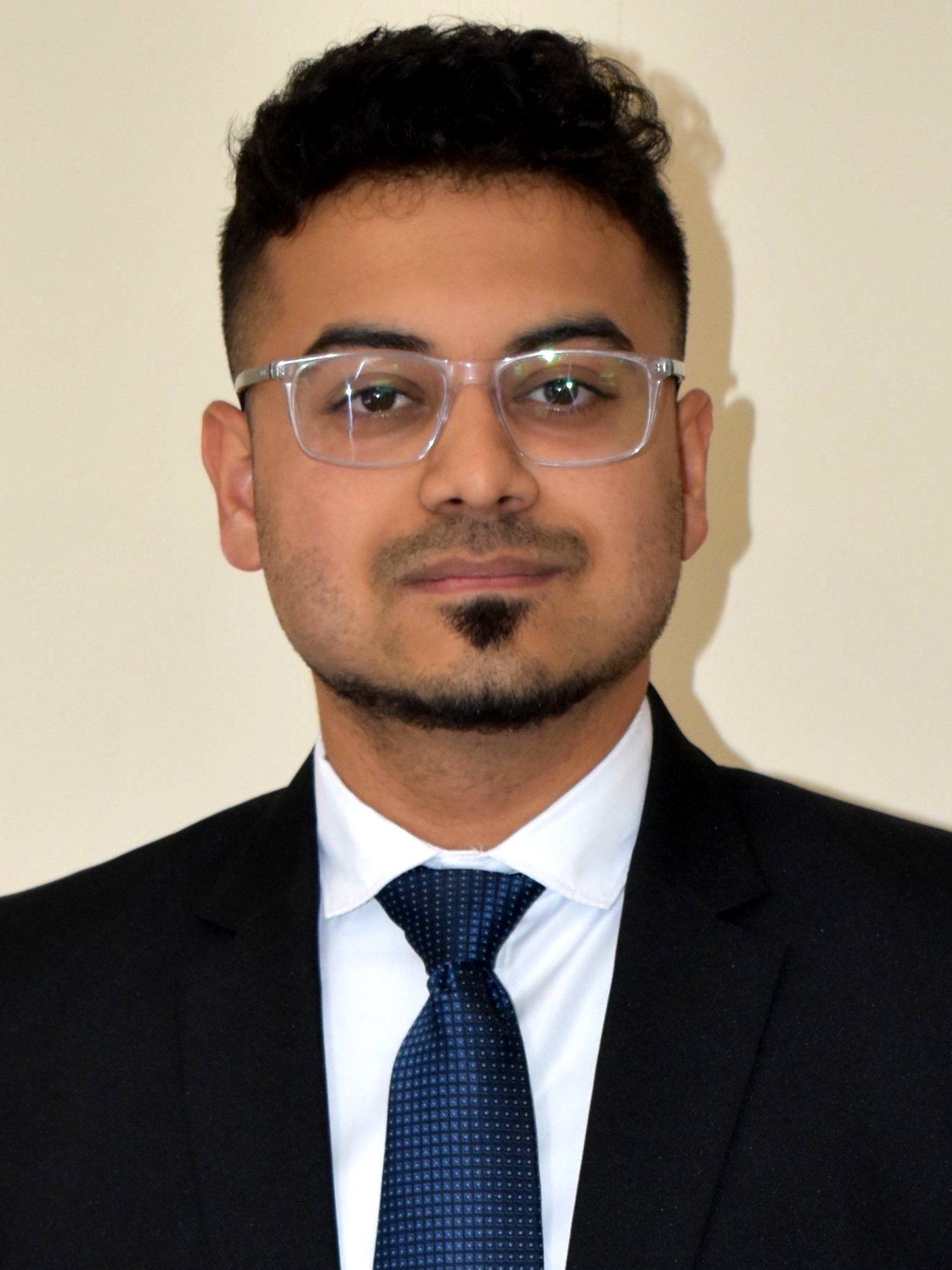}}]
    {Chinmaya Kaundanya}
    is a researcher at the School of Computing at Dublin City University. His research focuses on implementing Computer Vision Deep Neural Network model compression techniques and studying the challenges of deploying them on low-spec hardware platforms, especially for Micromobility applications. His research mainly includes different Neural Network model pruning and quantisation techniques and various low-spec Microcontroller Units.
\end{IEEEbiography}

\vspace{-2\baselineskip}

\begin{IEEEbiography}[{
    \includegraphics[width=1in,height=1.33in,clip,keepaspectratio]{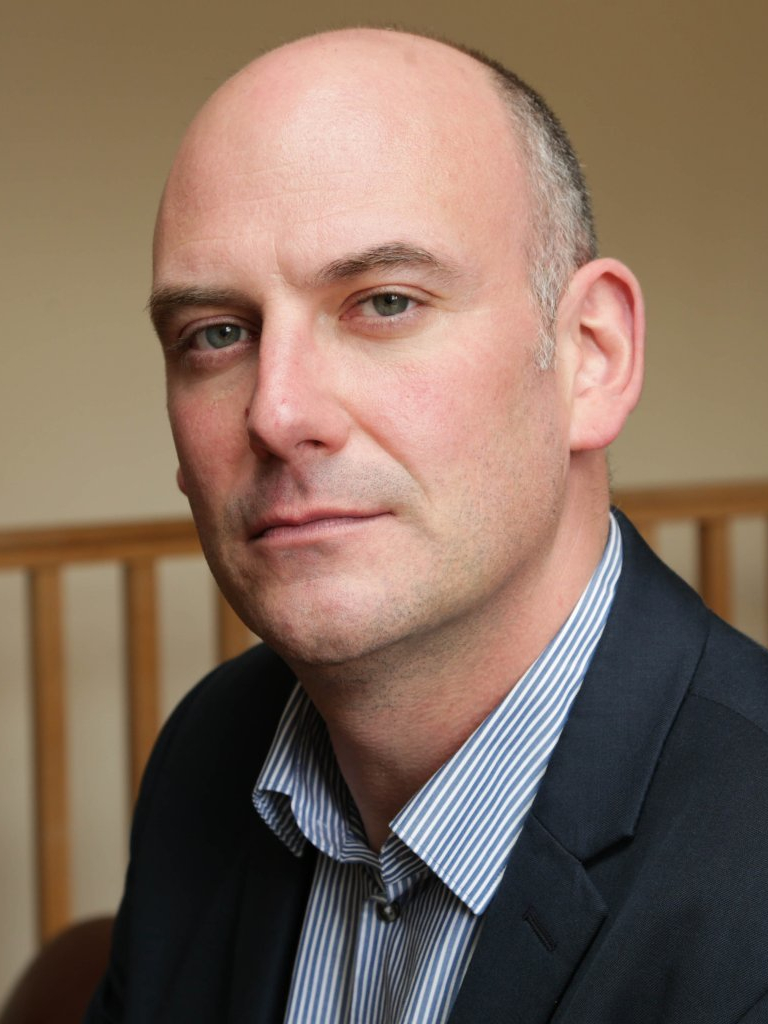}}]
    {Noel E. O’Connor (Senior Member, IEEE)}
    is a Full Professor at the School of Electronic Engineering, Dublin City University. He is Director of Rinn Artificial Intelligence, one of seven centres launched as part of Research Ireland’s enhanced national research network. His research focuses on multimedia content analysis using machine learning and AI, for applications in security/safety, autonomous vehicles, medical imaging, ambient assisted living, multimedia content-based retrieval, smart cities, precision agriculture, and environmental monitoring.
\end{IEEEbiography}

\vspace{-2\baselineskip}

\begin{IEEEbiography}[{
    \includegraphics[width=1in,height=1.33in,clip,keepaspectratio]{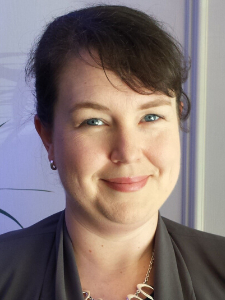}}]
    {Suzanne Little}
    is a Professor in the School of Computing at Dublin City University. She received the Ph.D. degree from the University of Queensland, Australia, in 2006. Her expertise is in video analysis, semantic search and data integration and her current research includes projects examining bias in deep learning models for computer vision, safety in micromobility and personal transportation, improved medical image analysis, biomedical data and water desalination.
    
\end{IEEEbiography}

\vspace{-2\baselineskip}

\begin{IEEEbiography}[{
    \includegraphics[width=1in,height=1.33in,clip,keepaspectratio]{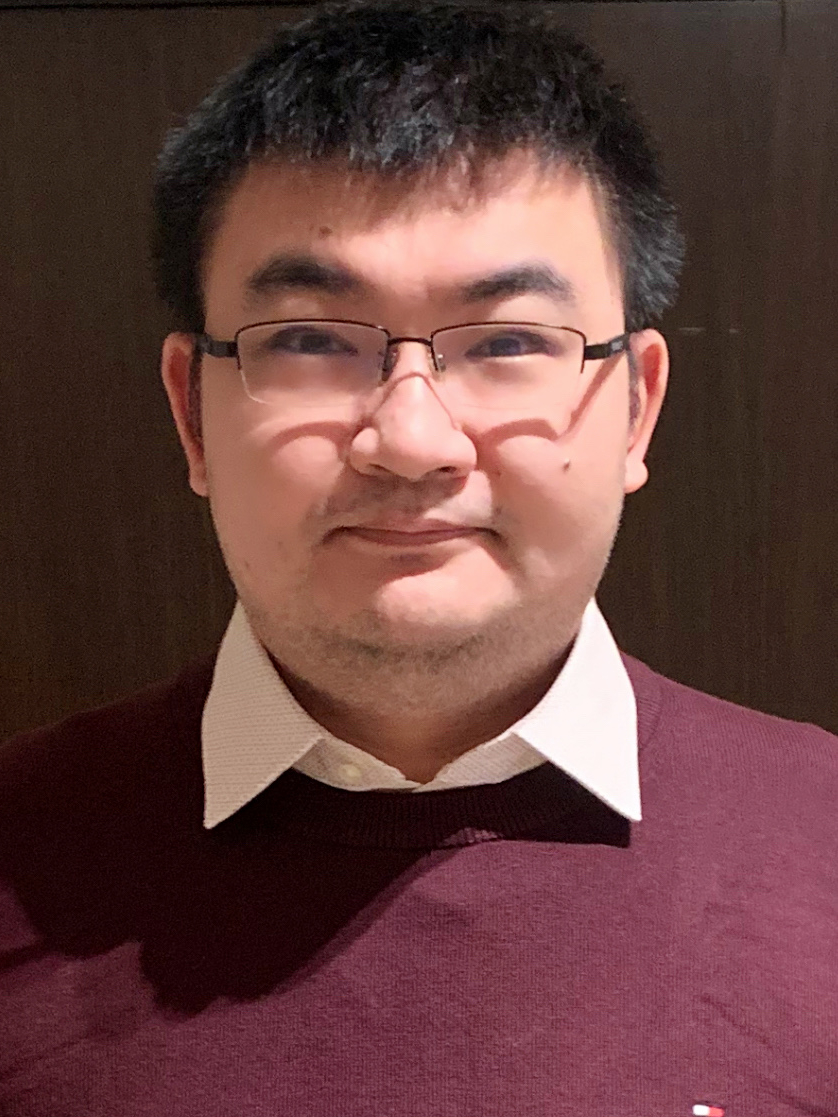}}]
    {Mingming Liu (Senior Member, IEEE)} 
    is an Associate Professor in the School of Electronic Engineering at Dublin City University. He received the B.Eng. degree in Electronic Engineering in 2011 and the Ph.D. degree in Control Engineering and Decision Science in 2015 from Maynooth University. He previously worked at University College Dublin and IBM Ireland. His research interests include control, optimisation, and machine learning for intelligent transportation, electric mobility, energy systems, and smart cities.
\end{IEEEbiography}

\vfill


 




\vfill

\end{document}